\RequirePackage{fix-cm}
\documentclass[twocolumn]{svjour3}          
\smartqed  

\usepackage{graphicx}
\usepackage[numbers,sort&compress]{natbib}
\usepackage{xcolor}
\usepackage{hyperref}
\hypersetup{
    colorlinks,
    linkcolor={black!50!black},
    citecolor={blue!50!black},
    urlcolor={blue!80!black}
}

\usepackage{lineno}
\usepackage{soul}
\usepackage{lscape}
\usepackage{booktabs}
\usepackage{amsmath,mathtools}
\usepackage[intoc]{nomencl}
\usepackage{url}
\usepackage{doi}
\usepackage[boxruled,vlined,linesnumbered]{algorithm2e}
\usepackage{amsfonts}
\usepackage{relsize}
\usepackage{commath}
\usepackage{float}
\usepackage{textcomp}
\usepackage{array}
\usepackage[nottoc,notlot,notlof]{tocbibind}
\usepackage{setspace}
\usepackage{mathptmx}

\usepackage{subcaption}
\graphicspath{{plots/}}

\usepackage{xcolor}
\definecolor{materialPurple}{HTML}{9C27B0}
\definecolor{materialTeal}{HTML}{009688}
\definecolor{materialGreen}{HTML}{4CAF50}
\definecolor{materialOrange}{HTML}{FF9800}
\definecolor{materialRed}{HTML}{FF0000}
\usepackage{setspace} 
\usepackage{lineno} 
\usepackage{comment}

\usepackage{csquotes}
\usepackage{bm}

\begin{document}

\title{Convolutional -- Recurrent Neural Network Proxy for Robust Optimization and Closed-Loop Reservoir Management
}
\author{Yong Do Kim \and Louis J.~Durlofsky}

\institute{Yong Do Kim \at
              Department of Energy Resources Engineering, \\
              Stanford University, Stanford CA 94305, USA \\
              \email{ykim07@stanford.edu}           
           \and
           Louis J. Durlofsky \at
              Department of Energy Resources Engineering, \\
              Stanford University, Stanford CA 94305, USA \\
              \email{lou@stanford.edu}
}

\date{Received: date / Accepted: date}
\maketitle

\begin{abstract} Production optimization under geological uncertainty is computationally expensive, as a large number of well control schedules must be evaluated over multiple geological realizations. In this work, a convolutional-recurrent neural network (CNN-RNN) proxy model is developed to predict well-by-well oil and water rates, for given time-varying well bottom-hole pressure (BHP) schedules, for each realization in an ensemble. This capability enables the estimation of the objective function and nonlinear constraint values required for robust optimization. The proxy model represents an extension of a recently developed long short-term memory (LSTM) RNN proxy designed to predict well rates for a single geomodel. A CNN is introduced here to processes permeability realizations, and this provides the initial states for the RNN. The CNN-RNN proxy is trained using simulation results for 300 different sets of BHP schedules and permeability realizations. We demonstrate proxy accuracy for oil-water flow through multiple realizations of 3D multi-Gaussian permeability models. The proxy is then incorporated into a closed-loop reservoir management (CLRM) workflow, where it is used with particle swarm optimization and a filter-based method for nonlinear constraint satisfaction. History matching is achieved using an adjoint-gradient-based procedure. The proxy model is shown to perform well in this setting for five different (synthetic) `true' models. Improved net present value along with constraint satisfaction and uncertainty reduction are observed with CLRM. For the robust production optimization steps, the proxy provides $O(100)$ runtime speedup over simulation-based optimization.


\end{abstract}

\section{Introduction}
\label{sec:1}
The goal in production optimization is to find optimal time-varying well settings such that an objective function is maximized or minimized. Because subsurface geology is inherently uncertain, it is important to include this uncertainty in the optimization. In this case, the objective function may be to maximize the expected net present value (NPV) of the operation, with the expectation computed as an average over a set of geological realizations. Inclusion of geological uncertainty therefore acts to increase computational demands, as each candidate well control schedule encountered during optimization must be evaluated over all geological realizations considered. This can become prohibitively expensive in practice, as it entails a large number of reservoir simulation runs, each of which may be computationally demanding.

In recent work \citep{Kim2021RNN}, we introduced a recurrent neural network (RNN)-based well control optimization procedure to maximize an objective function subject to nonlinear output constraints (examples of such constraints include maximum well or field-wide water injection/production rate). Our RNN procedure provides the oil and/or water rate time series for each well in a single (deterministic) model for a specified time-varying well bottom-hole pressure (BHP) schedule. Here we extend this methodology to predict well responses over multiple realizations, which enables us to perform robust optimization over an ensemble of geomodels. Specifically, by incorporating a convolutional neural network (CNN) with the RNN, our proxy model provides oil and water rate time series, for a specified BHP schedule, for every well in each realization in the ensemble. The proxy is incorporated into a closed-loop reservoir management (CLRM) workflow, in which we successively apply history matching followed by production optimization to maximize expected NPV.

A wide range of proxy models have been developed for production optimization (also called well control optimization) for a single deterministic model. These include treatments based on reduced-physics modeling \citep{moyner2015, rodriguez_torrado2015, deBrito2020}, reduced-order numerical models constructed using proper orthogonal decomposition \citep{Jansen2017UseOR}, machine-learning methods \citep{Guo2018RobustLP, Chen2020GlobalAL}, deep-learning procedures \citep{Zangl2006, Golzari2015DevelopmentOA, Kim2021RNN}, etc. We will not review this literature in detail, however, because our focus in this study is on the development of proxy models designed to treat geological uncertainty. Along these lines, Zhao et al.~\citep{ZHAO2020107192} and Zhang and Sheng~\citep{10.2118/206755-PA} used kriging as a statistical proxy in well control and hydraulic fracturing robust optimization problems. Petvipusi et al.~\citep{Petvipusit2014} applied an adaptive sparse grid interpolation surrogate to optimize CO$_2$ injection rates in carbon storage problems. Babaei and Pan~\citep{Babaei2016} assessed the performance of kriging, multivariate adaptive regression splines, and cubic radial basis function-based surrogates, and combinations of these techniques, in well control optimization problems under geological uncertainty.

Recent developments in machine learning and deep learning have inspired the implementation of proxy models for well control and well placement optimization under geological uncertainty. Guo and Reynolds~\citep{Guo2018RobustLP} applied support vector regression (SVR) as the forward model for robust production optimization. They compared two different approaches for constructing SVR-based proxies under geological uncertainty -- constructing a single proxy estimating average NPV across all realizations considered, and constructing separate proxies for each realization. They achieved better performance with the single-proxy approach. CNN-based proxy models have also been used to optimize well location. Specifically, Kim et al.~\citep{KIM2020107424} processed streamline time of flight and water injection and production well location maps using CNN to estimate NPV for each realization. Wang et al.~\citep{WANG2022109545} introduced a theory-guided CNN procedure to predict pressure maps, for multiple geomodels for given production well locations, under primary recovery. Nwachukwu et al.~\citep{Nwachukwu2018} used an extreme gradient boosting-based proxy to optimize well location and water-alternating-gas injection control parameters under geological uncertainty for CO$_2$ enhanced oil recovery. 

A proxy that can be applied for optimization over multiple geomodels will be especially useful in a closed-loop reservoir management setting. In CLRM, the model is updated (at a number of update/control steps) based on newly observed production data, and then re-optimized. This leads to uncertainty reduction (from the history matching step) and improved NPVs (from the robust well control optimization step). Traditional CLRM methods are expensive as they require many reservoir simulation runs for both steps \citep{Jansen2009,Sarma2006,10.2118/109805-PA}. Ensemble-based robust optimization \citep{10.2118/163657-PA,10.2118/173268-PA,Ramaswamy2020} has been used in CLRM, and this can lead to some improvement in efficiency. 
Recent studies have focused on further increasing computational efficiency in CLRM. Jiang et al.~\citep{Jiang2020}, for example, developed and applied a data-space inversion with variable controls (DSIVC) procedure to quickly generate posterior predictions and optimized controls. Deng and Pan~\citep{10.2118/200862-PA} replaced the reservoir simulator with a proxy that used an echo state network coupled with a water fractional flow relationship.

Both CNNs and RNNs are used in this work. CNNs are applicable for processing image data, while RNNs are used for time-series data. Both have been successfully applied in a great many applications \citep{SCHMIDHUBER201585, LeCun2015}. Recent developments relevant to our work entail the combination of CNN and RNN to handle image and time-series data together. Karpathy and Fei-Fei~\citep{Karpathy_2015_CVPR} combined CNNs over images and bidirectional RNNs over sentences to successfully generate natural language descriptions of images. In a subsurface flow setting, Tang et al.~\citep{TANG2020109456} developed a residual U-Net that used a convolutional long short-term memory (LSTM) RNN to predict dynamic pressure and saturation fields for new geomodels under fixed well control schedules. 
 
In this paper, we develop a CNN-RNN proxy to estimate time-varying well-by-well oil and water rates for each reservoir model in an ensemble. The geomodels, defined in terms of 3D permeability fields, along with the dynamic well BHP schedule, provide the inputs processed by the CNN and RNN, respectively. Vectors constructed by the CNN provide the initial long and short-term states for a sequence-to-sequence LSTM RNN \citep{Gers2000LearningTF}. The initial training of the network is based on simulation results from 300 full-order runs. In CLRM settings, the proxy is retrained after the ensemble of geomodels is updated in the history matching step. Detailed results demonstrating standalone proxy model performance, as well as its use in CLRM, will be presented. The system considered involves multiple 3D geomodels and oil recovery via water-flooding. As in \citep{Kim2021RNN}, well control optimization is achieved using the proxy in conjunction with particle swarm optimization (PSO), with a filter-based nonlinear constraint handling treatment \citep{Isebor2014}. History matching is accomplished using the randomized maximum likelihood method (RML) and gradient-based minimization. 

This paper proceeds as follows. In Section~\ref{sec:2}, we provide a detailed description of the CNN--RNN proxy developed in this work. In Section~\ref{sec:3}, the robust well control optimization problem is presented, and the incorporation of the proxy into a CLRM workflow is described. Detailed results for proxy performance, both standalone and in CLRM, are provided in Section~\ref{sec:4}. CLRM results for five different (synthetic) `true' models will be presented. We conclude in Section~\ref{sec:conclusion} with a summary and suggestions for future work in this area.

\section{CNN--RNN Proxy Model}
\label{sec:2}
In this section, we describe the key aspects of the CNN--RNN proxy developed in this work, including model architecture and the training process. TensorFlow \citep{tensorflow2015-whitepaper} is used to develop and train this proxy model.

\subsection{Network Architecture}
\label{sec:2.1}
Convolutional and recurrent neural network architectures are designed to process spatial data (CNN) and time-sequence data (RNN). In our previous study \citep{Kim2021RNN}, we developed an RNN-based proxy for well control optimization with a single deterministic geomodel. In this study, a CNN and an RNN  are combined to handle spatial geomodel and dynamic well control data together. 

\begin{figure*}[htbp!]\centering
\includegraphics[width=0.8\textwidth]{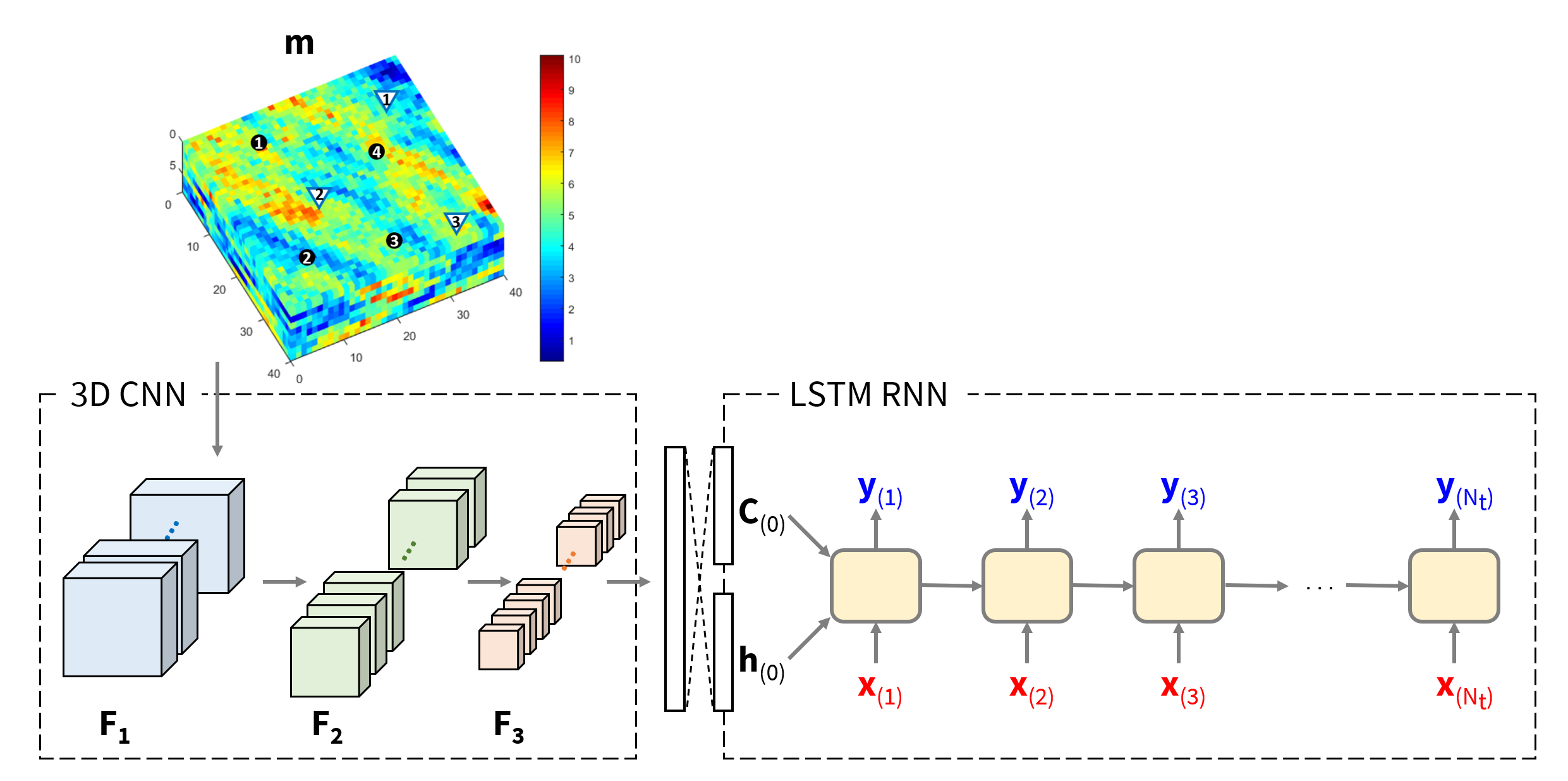}
\caption{CNN--RNN proxy structure. The CNN and RNN accept the geomodel $\textbf{m}$ (permeability field) and time-varying BHP profile $\textbf{x}_{(t)}$ as inputs. Proxy outputs are the time-varying well-by-well rates $\textbf{y}_{(t)}$. Here feature matrix $\textbf{F}_3$ from the CNN is flattened and then processed by a fully connected layer to generate initial long and short-term state vectors ($\textbf{c}_{(0)}$ and $\textbf{h}_{(0)}$) for the RNN.} \label{fig:proxy}
\end{figure*}

Figure~\ref{fig:proxy} depicts the structure of the CNN--RNN proxy developed in this study. The overall proxy accepts a particular 3D permeability realization and time-varying BHP profiles to estimate water injection rates for $n_I$ injectors, and oil and water production rates for $n_P$ producers, at each time step for the input geomodel. In the figure, the vectors $\textbf{x}_{(t)} \in \mathbb{R}^{N_{in}}$ and $\textbf{y}_{(t)} \in \mathbb{R}^{N_{out}}$ correspond to BHPs and well rates, respectively, at time step $t$. In this study, $N_{in} = n_I + n_P$ and $N_{out} = n_I + 2 \times n_P$. The total number of time steps is denoted $N_t$. We note that, in terms of notation, we use $N$ with a subscript to specify quantities associated with algorithmic settings, such as the number of neurons, number of inputs, etc., and $n$ with a subscript to specify parameters related to the reservoir simulation model, such as number of grid blocks and number of injectors or producers. 

Table~\ref{CNN-RNN-based proxy architecture details} provides the detailed architecture of the CNN--RNN proxy used in this study. Here, Conv, BN, Pool, and FC denote 3D convolutional layer, batch normalization \citep{Sergey2015}, 3D max-pooling layer, and fully connected layer, respectively. The parameter $N_{neu}$ is the number of neurons used in fully connected layers. We expand the 3D permeability field to a 4D tensor to enable processing with a 3D convolutional layer. The geomodel $\textbf{m}$, described in terms of an isotropic permeability value in every grid block, is of dimensions $n_x \times n_y \times n_z \times 1$, where $n_x$, $n_y$ and $n_z$ are the number of grid blocks in the $x$, $y$, and $z$ directions. If we were treating anisotropic models with different permeability components in each coordinate direction, the input dimensions would be $n_x \times n_y \times n_z \times 3$.

\begin{table*}[h]
\centering
\caption{CNN--RNN proxy architecture details.}
\label{CNN-RNN-based proxy architecture details}
\begin{tabular}{l c c}
 \hline
 Layer & Output size \\
 \hline
 Input 1 (Permeability) & $(n_x, n_y, n_z, 1)$ \\
 Conv 1, 4 filters of size 3 $\times$ 3 $\times$ 3, stride 1  & $(n_x, n_y, n_z, 4)$ \\
 BN 1 & $(n_x, n_y, n_z, 4)$  \\
 ReLU 1 & $(n_x, n_y, n_z, 4)$  \\
 Pool 1, pool size of 2 $\times$ 2 $\times$ 2 & $(n_x/2, n_y/2, n_z/2, 4)$\\
 Conv 2, 8 filters of size 3 $\times$ 3 $\times$  3, stride 1 & $(n_x/2, n_y/2, n_z/2, 8)$ \\
 BN 2 &  $(n_x/2, n_y/2, n_z/2, 8)$ \\
 ReLU 2 & $(n_x/2, n_y/2, n_z/2, 8)$ \\
 Pool 2, pool size of 2 $\times$ 2 $\times$ 2 & $(n_x/4, n_y/4, n_z/4, 8)$ \\
 Conv 3, 16 filters of size 3 $\times$ 3 $\times$  3, stride 1 & $(n_x/4, n_y/4, n_z/4, 16)$ \\ 
 BN 3 &  $(n_x/4, n_y/4, n_z/4, 16)$ \\
 ReLU 3 &  $(n_x/4, n_y/4, n_z/4, 16)$ \\
 Pool 3, pool size of 2 $\times$ 2 $\times$ 2 & $(n_x/8, n_y/8, n_z/8, 16)$ \\
 Flatten & $(n_x/8 \cdot n_y/8 \cdot n_z/8 , 1)$  \\
 FC 1  & $(N_{neu}, 1)$\\
 FC 2  & $(N_{neu}, 1)$\\
 Input 2 (BHP profile) & $(N_t, N_{in})$  \\ 
 LSTM & $(N_t, N_{neu})$ \\
 FC 3 & $(N_t, N_{out})$ \\
 \hline
 \end{tabular}
\end{table*}

The geomodel $\textbf{m}$ is first fed through the CNN. Each convolutional layer is followed by batch normalization, ReLU nonlinear activation, and a 3D max-pooling layer. Batch normalization is applied to stabilize the training process by shifting the mean to zero and normalizing each input.

The general CNN architecture, as used in this study, can be expressed as follows:
\begin{equation}
\textbf{F}_l = \mbox{Pool} \big[\mbox{ReLU}\big(\textbf{W}_l*\textbf{F}_{l-1} + \textbf{b}_l\big)\big]. 
\end{equation}
\noindent Here, $\textbf{F}_l$ and $\textbf{F}_{l-1}$ denote feature maps at layers $l$ and $l-1$ (note that $\textbf{F}_0=\textbf{m}$), $\textbf{W}_l$ designates a  3D kernel, and $\textbf{b}_l$ is the bias. The symbol $*$ denotes the convolution operation (i.e., the sum of the element-wise product of $\textbf{W}_l$ and the corresponding region of $\textbf{F}_{l-1}$). Pooling is the process of extracting a maximum or average value from each particular field. The 3D max-pooling layers, with $2 \times 2 \times 2$ pooling kernel, stride 2, and no padding, are applied in this study. The pooling layers reduce the dimensions of each feature map by a factor of 2.

As permeability input $\textbf{m}$ is processed through the CONV-BN-ReLU-Pool layers recursively, the dimensions of the feature map along the $x$, $y$, and $z$ directions decrease, while the total number of feature maps increases. After the last pooling layer (Pool~3), $\textbf{F}_3$ is flattened into a single vector. Two fully connected layers (FC~1 and FC~2) then process the reduced permeability vector to generate the initial long-term ($\textbf{c}_{(0)} \in \mathbb{R}^{N_{neu}}$) and short-term ($\textbf{h}_{(0)} \in \mathbb{R}^{N_{neu}}$) states for the LSTM. 

\begin{figure}[htbp!]\centering
\includegraphics[width=8cm]{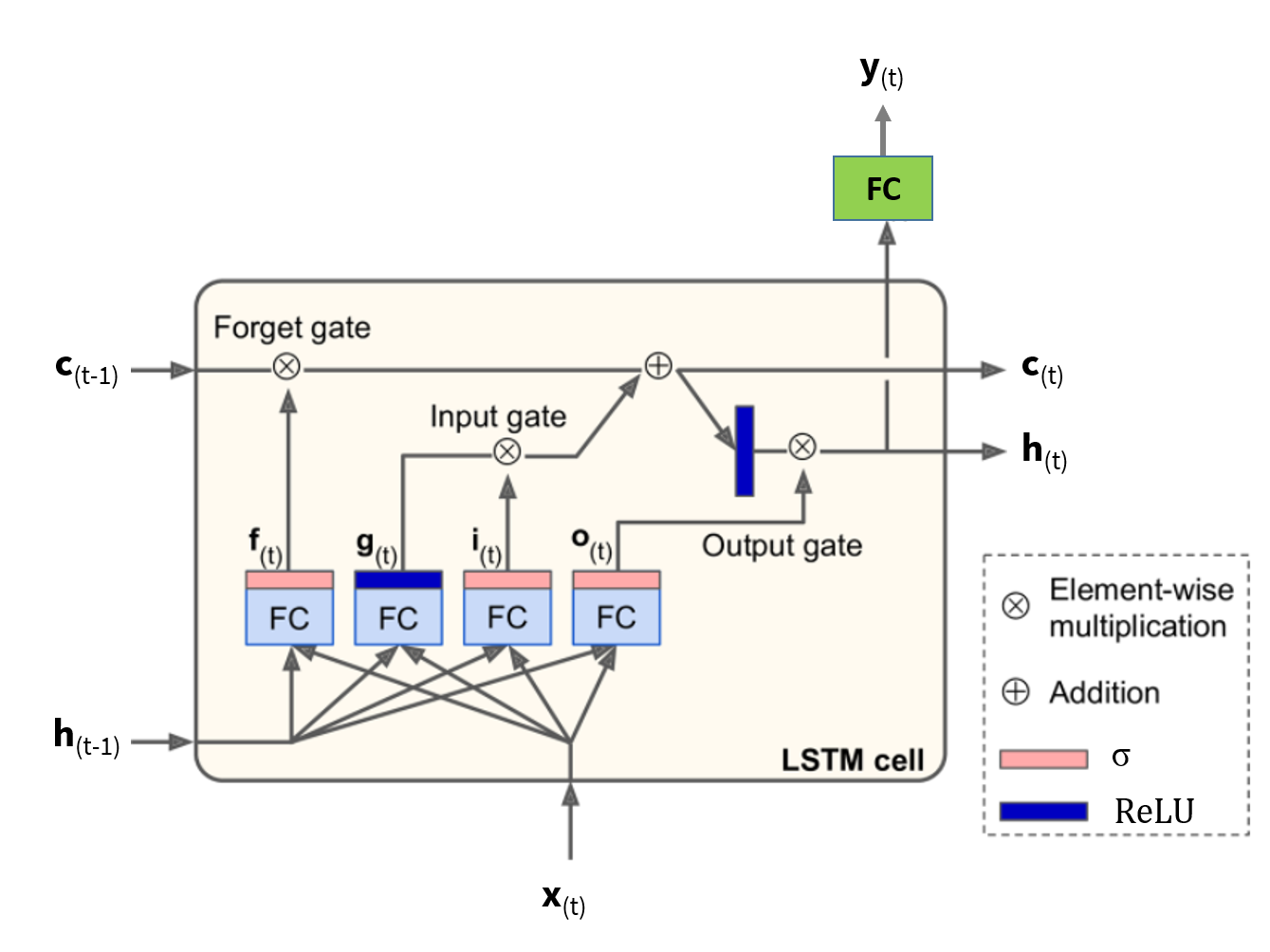}
\caption{LSTM cell (figure modified from \citep{Geron}).} \label{fig:LSTM}
\end{figure}

We utilize LSTM cells to process time-sequence input and output data. LSTM has been shown to perform well in time-series problems as it is able to capture both long-term and short-term dependencies in the data. This is achieved by tracking long-term ($\textbf{c}_{(t)} \in \mathbb{R}^{N_{neu}}$) and short-term ($\textbf{h}_{(t)} \in \mathbb{R}^{N_{neu}}$) states. The short-term state carries information from the previous time step, while the long-term state contains key information from time steps prior to $t$. We use the same LSTM RNN (Figure~\ref{fig:LSTM}) as was used in \citep{Kim2021RNN}. Here, however, static permeability information from the CNN provides the initial long-term and short-term states, $\textbf{c}_{(0)}$ and $\textbf{h}_{(0)}$. 

We now briefly describe the LSTM RNN (for full details, please see \citep{Kim2021RNN}). At each time step $t$, the input vector $\textbf{x}_{(t)}$ and the short-term state $\textbf{h}_{(t-1)}$ are processed by the four different fully connected layers according to the following expression
\begin{align}
\label{eqn:LSTMComputation}
\textbf{z}_{(t)} &= f\big(\textbf{W}_{xz}^T\cdot\textbf{x}_{(t)} + \textbf{W}_{hz}^T\cdot\textbf{h}_{(t-1)} + \textbf{b}_z\big).
\end{align}
\noindent Here $\textbf{z}$ represents each of the quantities $\textbf{f}$, $\textbf{i}$, $\textbf{o}$, and $\textbf{g}$, where $\textbf{f}_{(t)}$, $\textbf{i}_{(t)}$, and $\textbf{o}_{(t)}$ control the forget gate, the input gate, and the output gate, respectively. The main layer outputs a vector of candidate values, $\textbf{g}_{(t)}$, that can be added to the long term state. 

The function $f$ corresponds to a nonlinear activation, specifically ReLU (for $\textbf{g}$) or sigmoid (for $\textbf{f}$, $\textbf{i}$, and $\textbf{o}$). The weight matrices $\textbf{W}_{xz}$ and $\textbf{W}_{hz}$ ($z=g,f,i,o$) process $\textbf{x}_{(t)}$ and the short-term state $\textbf{h}_{(t-1)}$. The vectors $\textbf{b}_{z}$ ($z=g,f,i,o$) are the bias terms for each of the four layers. The initial short-term state $\textbf{h}_{(0)}$ is fed to Equation~\eqref{eqn:LSTMComputation} when the RNN time step $t$ is 1.

The long-term state $\textbf{c}_{(t)}$ is updated by the forget gate and the input gate through application of
\begin{equation}
\label{c_update}
\textbf{c}_{(t)} =  \textbf{f}_{(t)} \otimes \textbf{c}_{(t-1)}+ \textbf{i}_{(t)} \otimes \textbf{g}_{(t)},
\end{equation}
\noindent where $\otimes$ indicates element-wise multiplication. The initial long-term state $\textbf{c}_{(0)}$ is used in Equation~\eqref{c_update} when $t=1$. Element-wise multiplication between the updated long-term state $\textbf{c}_{(t)}$ and the output vector $\textbf{o}_{(t)}$  is performed in the output gate to provide the updated short-term state $\textbf{h}_{(t)}$, i.e., 
\begin{equation}
\label{h_update}
\textbf{h}_{(t)} = \textbf{o}_{(t)} \otimes f(\textbf{c}_{(t)}). 
\end{equation}

The long-term and short-term state vectors $\textbf{c}_{(t)}$ and $\textbf{h}_{(t)}$ are updated recurrently using Equations~\eqref{eqn:LSTMComputation}, \eqref{c_update}, and \eqref{h_update}. As explained earlier, the CNN constructs the initial states from the input geomodel \textbf{m}. This enables static permeability information, along with dynamic well control schedules and well rate information saved in $\textbf{c}_{(t)}$ and $\textbf{h}_{(t)}$, to propagate through the LSTM RNN. 

\subsection{Training Procedure}
\label{sec:2.2}
We now describe how the CNN--RNN proxy is trained to predict well rates for multiple realizations in an ensemble and a given BHP profile. The training is performed for a set of $n_r$ permeability realizations. In a closed-loop setting, these are prior geomodels in the initial cycle, and history matched models in later cycles. We generate $n_r \times N_B$ different BHP profiles by randomly sampling from a uniform distribution within specified bounds. Each realization in the ensemble is then paired with $N_B$ BHP profiles, and high-fidelity simulation is performed for the $n_r \times N_{B}$ cases.

In our setup, the BHP of each well is shifted every $T_{cs}$ days, a total of $n_{cs}$ times over the full simulation time frame ($n_{cs}$ is the number of control steps, each of the same duration). We use $t$ and $T$ to denote the RNN time step and the actual time in days. Denoting $T_{N_t}$ as the final time, we thus have $T_{cs}=T_{N_t}/n_{cs}$. The total number of BHP control variables is $(n_I + n_P) \times n_{cs}$.

All high-fidelity simulations are performed using Stanford's Automatic Differentiation-based General Purpose Research Simulator, AD-GPRS \citep{zhouThesis}. The CNN--RNN proxy predicts well rates every 30~days from day~0 to the final simulation time $T_{N_t}$ (here $T_{N_t}=900$~days). Thus the RNN time step $t$ corresponds to simulation time $30~\times~(t-1)$. The $n_r~\times~N_{B}$ simulation runs are divided into a training set containing results for $n_r~\times~N_{B,train}$ runs and a test (validation) set, containing results for $n_r~\times~N_{B,test}$ runs. 

During training, the loss function (discussed later in this section) is minimized by adjusting the parameters for the CNN, the LSTM RNN, and the three fully connected layers. We apply the same error metric $E$ as in \citep{Kim2021RNN} for training and evaluation. The error for fluid phase $f$ ($f = o$ for oil and $w$ for water) from production or injection well $j$, for sample $i$, is defined as

\begin{equation}
\label{fluidloss}
e_{f}^{WT,j,i} = \sum_{t=2}^{N_t} \frac{\left|{q_{sim,f}^{WT,j,i}(T_t) - q_{proxy,f}^{WT,j,i}(T_t)} \right|}{ q_{sim,f}^{WT,j,i}(T_t) + \alpha_f^{WT}}.
\end{equation}

\noindent Here, $q_f^{WT,j,i}(T)$ denotes the fluid rate for production or injection well $j$ in sample $i$ at time $T$. The superscript $WT$ denotes well type ($WT = P$ for producer and $I$ for injector). The subscripts $sim$ and $proxy$ denote the high-fidelity simulation and CNN--RNN proxy results. With this notation, $e_{o}^{P,j,i}$, for example, denotes the oil production rate error for production well $j$ in sample $i$. The constant $\alpha_f^{WT}$ enters to prevent $e_{f}^{WT,j,i}$ from becoming very large when $q_{sim,f}^{WT,j,i}(T)$ is close to 0. This is only required for water production (since these rates are zero prior to water breakthrough), in which case we set $\alpha_w^{P}= 20$~m$^3$/d. For oil production and water injection, we set other $\alpha_o^{P}=\alpha_w^{I}=0$. We define the overall time-average error for sample $i$ as 
\begin{equation} \label{Ei}
E^{i}  = \frac{1}{n_{I}+2n_{P}} \times \Bigg(\sum_{j=1}^{n_{I}}e_{w}^{I,j,i} + \sum_{j=1}^{n_{P}}e_{o}^{P,j,i} + \sum_{j=1}^{n_{P}}e_{w}^{P,j,i}\Bigg),
\end{equation}
\noindent and the error for the entire training set as
\begin{equation} \label{E}
E = \frac{1}{n_r \times N_{B,train}} \times \Big(\sum_{i=1}^{n_r \times N_{B,train}}E^{i}\Big).
\end{equation}

Consistent with the approach in \citep{Kim2021RNN}, the loss function $L$ differs slightly from $E$ in Equation~\ref{E} in that the sum in Equation~\ref{fluidloss} starts at $t=1$. Please see \citep{Kim2021RNN} for further discussion. Training is performed to minimize $L$, though it is terminated when $E$ reaches a specified tolerance. 

In the initial CNN--RNN proxy training, for the 3D case considered here, we consider a total of 20 realizations in the ensemble ($n_r=20$). In \citep{Kim2021RNN}, we used 256 training simulation runs to train the RNN-based proxy (for a single realization). Here we specify $N_{B,train}=15$ and $N_{B,test}=10$, which results in 500 total simulation runs. We use the Adam optimizer \citep{Kingma2014} to minimize $L$. The initial learning rate is 0.001, which s is decreased to 0.0001 as $E$ decreases. Training is terminated when $E<0.05$. The batch gradient method is used for the gradient calculation. The size of batch size is $n_r \times N_{B,train}$. The total number of trainable parameters in the proxy is 333,523 for the 3D example considered in this work. For our case, the initial training process requires about 1.5~hours using a Nvidia Tesla V100 GPU. 

In the CLRM setting considered here the proxy is retrained four times, after each of the history matching steps. In the retraining process, we prescribe $N_{B,train}$ and the learning rate to be 10 and 0.0001. We set $N_{B,test}=3$ for 10 of the realizations and $N_{B,test}=2$ for the other 10 realizations. The retraining steps require from about 2~minutes to about 25~minutes, with shorter training times at later CLRM cycles (for reasons explained later).

\section{Closed-loop Reservoir Management using CNN--RNN Proxy}
\label{sec:3}
In this section, we first describe the robust production optimization problem considered in this work. PSO, with a filter-based treatment for nonlinear constraints, is then briefly discussed. We then present the PCA representation of the geomodels, the randomized maximum likelihood (RML) procedure used for history matching, and the overall CLRM workflow with the CNN--RNN proxy.

\subsection{Robust Production Optimization Problem}
Production optimization entails the determination of time-varying well control variables (BHPs in this case) that maximize an objective function, often taken to be cumulative oil produced or NPV. When geological uncertainty is considered, robust optimization, in which the expected value of the objective function is evaluated over multiple geological realizations, is applied. In this study, NPVs for the full set of $n_r$ realizations are evaluated first. The realizations corresponding to the upper and lower 10\% (in terms of NPV) are then excluded from the objective function calculation and nonlinear constraint treatment. This step is not essential, but we proceed in this way to eliminate extreme cases, as we have seen that these cases can have a disproportionate effect in terms of nonlinear constraint satisfaction (we require all geomodels used in the NPV calculation to satisfy all constraints). More specifically, in order to satisfy feasibility, these extreme cases may act to significantly limit the search space, thus yielding overly conservative solutions. 

A range of nonlinear constraint treatments have been considered in previous studies involving production optimization under uncertainty. For example, Chen et al.~\citep{10.2118/141314-PA} required nonlinear constraints to be satisfied in every realization. Hansen et al.~\citep{HANSSEN201562} adjusted risk parameters to control the probability of constraint violation, though some amount of violation was tolerated. Aliyev and Durlofsky~\citep{Aliyev2017} penalized realizations (in terms of objective function value) that did not satisfy the nonlinear constraints, though again, some violation was allowed. The CNN--RNN proxy developed in this work is compatible with different approaches for handling nonlinear constraints, as it provides well-by-well rates as a function of time.

In this study, the expected NPV is calculated from the well-rate time series data as follows: 
\begin{equation}
\label{NPV}
\begin{aligned}
& E[NPV] = \\
& \frac{1}{n_{ro}}\sum_{s=1}^{n_{ro}}\int_{T_1}^{T_{N_t}} \frac{\big[P_{p,o}Q^s_{p,o}(T) - C_{p,w}Q^s_{p,w}(T) - C_{i,w}Q^s_{i,w}(T)\big]}{(1+\gamma)^\frac{T}{365}}dT.
\end{aligned}
\end{equation}
\noindent Here, $n_{ro} = 0.8n_r$ is the number of realizations considered in the NPV calculation, $s=1, \dots, n_{ro}$ denotes a particular realization, $P_{p,o}$ is the price of oil, $C_{p,w}$ and $C_{i,w}$ are the costs for produced-water handling and water injection, with prices and costs in USD/stock-tank barrel (STB). The quantities $Q^s_{p,o}$, $Q^s_{p,w}$, and $Q^s_{i,w}$ are field-wide oil production, water production, and water injection rates, in STB/d, for realization $s$. The yearly discount rate is denoted by $\gamma$.

The constrained robust optimization problem is written as: 
\begin{equation}
\label{Optimization}
\begin{aligned}
\underset{\textbf{u} \in \textit{U}}{\mathrm{min}} \quad &  \textit{J}(\textbf{x}, \textbf{u}) = -E[NPV],\\
\textrm{subject to} \quad & \textbf{g}(\textbf{x}, \textbf{u}) = \textbf{0},\quad\textbf{x}|_{T=0} = \textbf{x}_0,\\
&\textbf{u}^L \leq \textbf{u} \leq \textbf{u}^U, \quad \textbf{c}(\textbf{x}, \textbf{u}) \leq \textbf{0}.\\
\end{aligned}
\end{equation}
\noindent Here, $J=-E[NPV]$ is the objective function to be minimized. The vector $\textbf{x}$ represents the dynamic reservoir state variables -- in the oil-water system considered in this study, these are pressure and saturation in each grid block at each time step. Initial reservoir states are defined as $\textbf{x}_0$. The vector $\textbf{u} \in U \subset \mathbb{R}^{n_v}$, where $n_v=(n_I + n_P)~\times~n_{cs}$, contains the well BHP optimization variables. The optimization problem is subject to the reservoir simulation equations $\textbf{g} = \textbf{0}$, bound constraints [$\textbf{u}^U, \textbf{u}^L$] (which act to define the space $U$), and nonlinear output constraints $\textbf{c} \in \mathbb{R}^{n_c}$. Nonlinear output constraints must be satisfied for the $n_{ro}$ realizations included in the objective function calculation.

\subsection{PSO with Nonlinear Constraints}
\label{sec:3.2}
As in \citep{Kim2021RNN}, we apply the CNN--RNN proxy for function and nonlinear constraint evaluations within PSO. A key difference here is that multiple realizations are considered. 
The following description of PSO is brief -- please see \citep{Kim2021RNN} for more details.  

In PSO, a set of $N_s$ particles, referred to as the swarm, moves through the search space from  iteration to iteration. The optimization proceeds until a termination criterion is met, here specified to be the number of PSO iterations. At each iteration, the particle locations are updated through application of 
\begin{equation}
\label{PSOposition}
\textbf{u}_{j,k+1} = \textbf{u}_{j,k} + \textbf{v}_{j,k+1}\qquad   \forall  j \in \{1,2,...,N_s\},
\end{equation}
\noindent where $\textbf{u}_{j,k}$ and $\textbf{u}_{j,k+1}$ are the locations of particle $\textit{j}$ at the end of iterations $\textit{k}$ and ${k+1}$. The PSO velocity, $\textbf{v}_{j,k+1}$, is a weighted sum of three contributions, referred to as the inertial, cognitive, and social components. The inertial term is simply the particle velocity from the previous iteration -- this acts to maintain some continuity in the search. The cognitive term moves the particle toward the best position the particle itself has experienced thus far in the optimization, while the social term directs the particle toward the position of the best particle in its interaction set (referred to as its neighborhood). The weights for the inertial, cognitive, and social terms are set to 0.729, 1.494, and 1.494, though the cognitive and social terms are multiplied by random values to add a stochastic component to the search. A random neighborhood topology is used for the social term. The detailed velocity expressions are provided in \citep{Kim2021RNN}.

The filter method developed by Isebor et al.~\cite{Isebor2014} is utilized to enforce feasibility. We represent aggregate constraint violations in terms of a function $h$. In the optimizations, minimizing $h$ takes precedence over minimizing $J$ as long as $h>0$, which indicates infeasibility. In the case of a single realization, the aggregate constraint violation function for particle $j$ at iteration $k$, for a maximum-bound nonlinear constraint, is given by \citep{Kim2021RNN} 
\begin{equation}
\label{constraint}
\begin{aligned}
&\textit{h}_{j,k}(\textbf{u}_{j,k}) = \sum_{l=1}^{n_c} \bar{c}_{j,k,l}(\textbf{u}_{j,k}),\\
&\textrm{where,}\\
&\bar{c}_{j,k,l}(\textbf{u}_{j,k}) =
\begin{cases}
     0 & \text{if}\ M_{k,l}(\textbf{u}_{j,k}) \leq c_{l,cons} \\
     \scalebox{1.25}{$\frac{c_{j,k,l}(\textbf{u}_{j,k})-c_{l,cons}}{M_{k,l}(\textbf{u}_{j,k})-c_{l,cons}}$} & \text{if}\ M_{k,l}(\textbf{u}_{j,k}) > c_{l,cons}.
    \end{cases}
\end{aligned}
\end{equation}
\noindent Here $\bar{c}_{j,k,l}(\textbf{u}_{j,k})$ is the normalized constraint violation for constraint $l$ and ${c_{j,k,l}}(\textbf{u}_{j,k})$ denotes the maximum value of constraint $l$ observed over all time steps. The value $M_{k,l}(\textbf{u}_{j,k})$ is the maximum value of ${c_{j,k,l}}(\textbf{u}_{j,k})$ over the entire PSO swarm at iteration $k$. The (specified) maximum bound for constraint $l$ is $c_{l,cons}$.

For the case of $n_{ro}$ realizations, ${c_{j,k,l}}(\textbf{u}_{j,k})$ generalizes to
\begin{equation}
\label{cjkl}
c_{j,k,l}(\textbf{u}_{j,k}) = \max_{i \in (1,n_{ro})} \left( c_{j,k,l}^{i}(\textbf{u}_{j,k}) \right),
\end{equation}
\noindent where $c_{j,k,l}^{i}(\textbf{u}_{j,k})$ is the maximum value of constraint $l$ observed over all time steps for realization $i$. The value $M_{k,l}(\textbf{u}_{j,k})$ is again the maximum of ${c_{j,k,l}}(\textbf{u}_{j,k})$ over the entire PSO swarm at iteration $k$. In this case, given the definition of $c_{j,k,l}^{i}$ in Equation~\eqref{cjkl}, $M_{k,l}$ is however the maximum over the swarm and over all $n_{ro}$ realizations. Different (but analogous) expressions are used for minimum-bound nonlinear constraints. 

For every candidate PSO solution, $n_{ro}$ realizations must be selected from the full set of $n_r$ realizations for function and constraint-violation evaluations. Recall that $n_{ro}=0.8n_r$, with the realizations corresponding to the upper and lower 10\% of NPVs excluded. This selection is performed for each particle at every PSO iteration. The selected realizations can thus differ between PSO particles, which is not a problem for our treatment. Very similar sets of realizations are typically used between particles, however, as extreme cases tend to correspond to the same realizations.

\subsection{Geomodel Parameterization and History Matching}

Geological realizations are parameterized in this study using principal component analysis (PCA). Such representations have been applied in a number of previous studies -- see \citep{Sarma2006} for discussion. PCA representations are suitable for use with multi-Gaussian models, which are the types of models considered in this work. For non-Gaussian models, e.g., channelized systems, deep-learning-based approaches such as the recent 3D CNN-PCA procedure \citep{LIU2021104676} are applicable.

To construct the PCA representation, we follow the approach described in \citep{LIU2021104676}. The goal is to represent geomodels $\mathbf{m} \in \mathbb{R}^{n_g}$, where $n_g$ is the number of grid blocks in the geomodel ($n_g=n_x n_y n_z$), in terms of a low-dimensional latent variable $\boldsymbol{\xi}~\in~\mathbb{R}^{l}$, with $l \ll n_g$. To achieve this, we first generate $n_{rp}$ random prior models conditioned to hard data at well locations. These prior models can be generated using geomodeling software; here we use the Stanford Geostatistical Modeling Software, SGeMS \citep{SGeMs}.

A centered data matrix $Y \in \mathbb{R}^{n_g \times n_{rp}}$ is then constructed
\begin{equation}
Y= 
\frac{1}{\sqrt{n_{rp}-1}}\big[\mathbf{m}_1-\mathbf{\bar m}\quad \mathbf{m}_2-\mathbf{\bar m}\quad ...\quad \mathbf{m}_{rp}-\mathbf{\bar m}\big],
\end{equation}
\noindent where $\mathbf{\bar m}$ denotes the prior mean. A singular value decomposition of $Y$ is performed, which allows us to write $Y=U \Sigma V^T$, where 
$U$ and $V$ are the left and right singular matrices and $\Sigma$ is a diagonal matrix containing the singular values. The PCA representation allows us to generate new realizations $\textbf{m}$ through use of 
\begin{equation} \label{eq:pca}
\mathbf{m} = U_l {\Sigma}_l \boldsymbol{\xi} + \mathbf{\bar m},
\end{equation}
\noindent where ${U}_l \in \mathbb{R}^{n_g \times l}$ and ${\Sigma}_l \in \mathbb{R}^{l \times l}$. The value of $l$ is selected to preserve 85\% of the total `energy' in the PCA representation (energy here involves the squared singular values in $\Sigma$; see \citep{LIU2021104676} for discussion). For the models considered in this work, this results in $l=219$. 

To generate random realizations, the components of $\boldsymbol{\xi}$ are sampled independently from a standard normal distribution. In the context of history matching, these components are determined such that the mismatch between simulation results and observed data is minimized. 

The history matching procedure applied in this work is the randomized maximum likelihood method (RML) with gradient-based minimization. The implementation is as described in \citep{Vo2015}. The minimization problem is expressed as
\begin{equation}
\label{RML_HM}
\begin{aligned}
&\boldsymbol{\xi}_{rml} = \\
&\arg\min_{\boldsymbol{\xi}}\big[(\mathbf{d}_{sim}-\mathbf{d}_{obs}^*)^TC_d^{-1}(\mathbf{d}_{sim}-\mathbf{d}_{obs}^*) + (\boldsymbol{\xi}-\boldsymbol{\xi}^*)^T(\boldsymbol{\xi}-\boldsymbol{\xi}^*)\big],
\end{aligned}
\end{equation}
\noindent where $\mathbf{d}_{sim} = \mathbf{d}_{sim}\mathbf{(m(\boldsymbol{\xi}))}$ denotes simulated flow data, $\mathbf{d}_{obs}^*$ represents observed data with noise, with noise sampled from $N(\mathbf{0},C_d)$, where $C_d$ is the covariance matrix of the measurement error. The vector $\mathbf{\boldsymbol{\xi}}^*$ denotes a prior realization of $\boldsymbol{\xi}$ sampled from $N({\bf 0},I)$. The  $(\boldsymbol{\xi}-\boldsymbol{\xi}^*)^T(\boldsymbol{\xi}-\boldsymbol{\xi}^*)$ term is essentially a regularization that acts to keep $\boldsymbol{\xi}_{rml}$ somewhat `close' to the prior realization $\boldsymbol{\xi}^*$. 

A total of $n_r$ posterior realizations of $\boldsymbol{\xi}_{rml}$ (and thus $\mathbf{m}$ through application of Equation~\eqref{eq:pca}) are generated by solving the minimization problem $n_r$ times, using different $\mathbf{d}^*$ and $\boldsymbol{\xi}^*$ for each run. This minimization is achieved using an adjoint-gradient-based approach, with the adjoint constructed by ADGPRS. For a more detailed description of the overall history matching procedure, please refer to \citep{Vo2015}. 

The history matching computations involve high-fidelity simulations -- a proxy model is not used in this step. Because the computational demand associated with the optimizations is much larger than that for history matching (computational requirements will be provided later), substantial overall savings are still achieved. We note finally that additional savings could be obtained by using a surrogate model for history matching, such as the recurrent R-U-Net described in \citep{TANG2021113636}. 

\subsection{Incorporation of CNN--RNN Proxy into CLRM}

\begin{figure}[htbp!]\centering
\includegraphics[width=8cm]{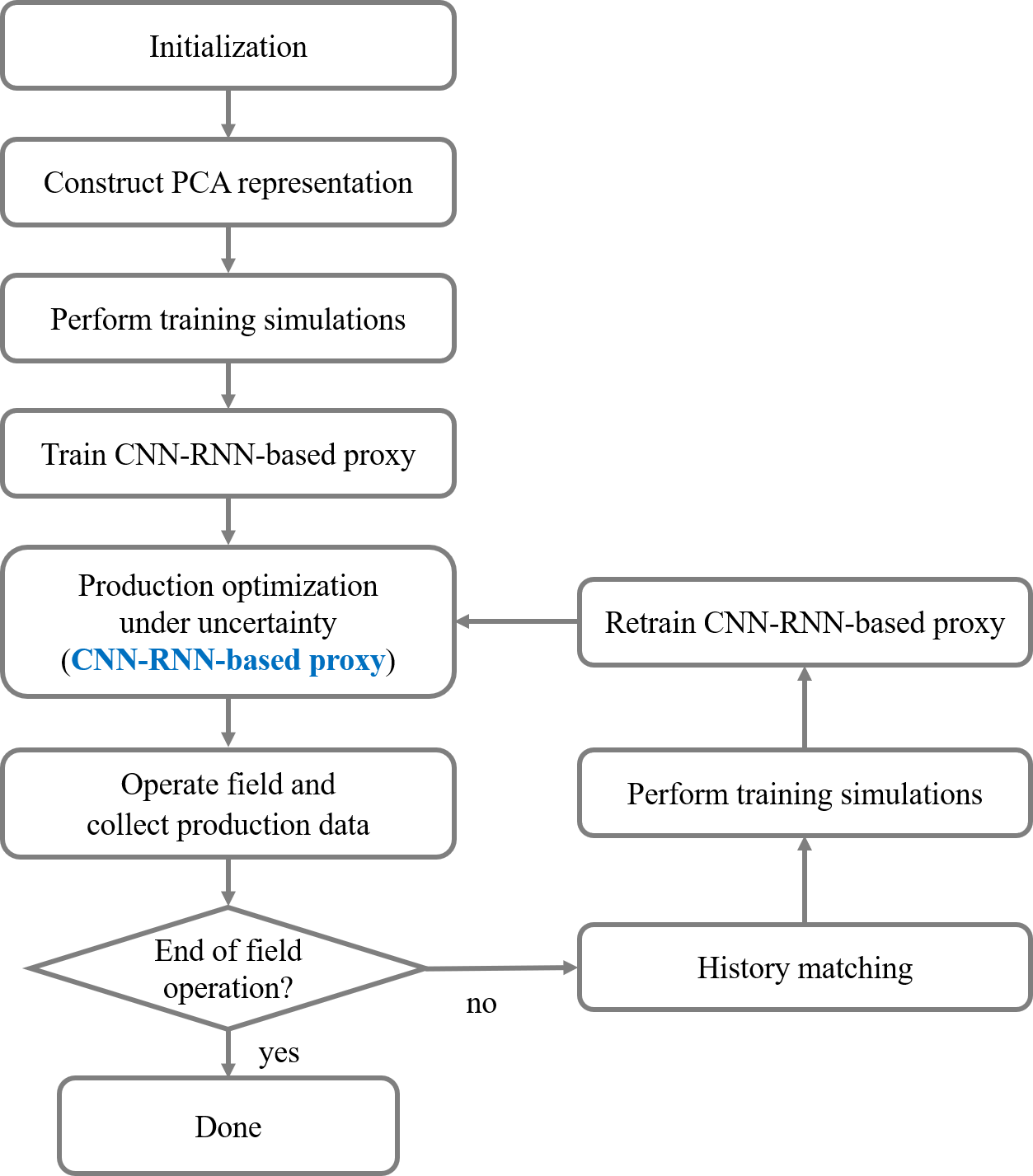}
\caption{Flowchart of the closed-loop reservoir management procedure using the CNN-RNN proxy for robust production optimization.} \label{fig:flowchartCNN-RNNCLRM}
\end{figure}

A flowchart for the full CLRM with the CNN--RNN proxy is shown in Figure~\ref{fig:flowchartCNN-RNNCLRM}. We first construct the PCA representation and train the initial proxy model using simulation results for pairs of permeability realizations and BHP profiles. We then perform robust production optimization using the CNN--RNN proxy for function and constraint evaluations in PSO. The proxy provides estimates of oil and water production and injection rates (in time) for specified BHP schedules for $n_{r}$ realizations in the ensemble. Speedup in the CLRM is achieved as a great number of different well control settings, for many realizations, are evaluated using the fast proxy rather than high-fidelity simulation.

In the first CLRM cycle ($n_{cyc} = 1$, where $n_{cyc}$ is the CLRM cycle number), the first $n_r$ realizations from the set of $n_{rp}$ realizations (applied to construct the PCA representation) are used as initial prior models. All realizations are conditioned to well data. Robust production optimization provides optimal well BHP schedules $\textbf{u}^1$ from day~0 to $T_{N_t}$ (superscript~1 denotes optimal BHPs from CLRM cycle~1). The field is operated based on $\textbf{u}^1$ until day~$T_{cs}$ is reached. Well production and injection rate data from day~0 to $T_{cs}$ are collected. The first CLRM history matching (using RML and adjoint-gradient minimization) is performed at day~$T_{cs}$, using all data observed up to this point.  

We next perform retraining with the new (history matched) set of $n_r$ realizations. Random BHP profiles within the allowable bounds are generated as described earlier, except now BHPs from day~0 to $T_{cs}$ are fixed at their values from the first cycle optimization (since we are now proceeding forward from $T_{cs}$). Simulations are performed to provide the quantities needed for retraining. For the 3D example considered in this study, retraining is performed using 200 new simulation runs ($n_r = 20$ and $N_{B,train}=10$). In general, training time decreases as we proceed through CLRM cycles, both because the geomodels become more similar (due to history matching over progressively longer periods) and because the BHP profiles differ from one another over fewer cycles (since there are progressively fewer future cycles). As noted earlier, retraining requires from $\sim$2--25~minutes.

The robust optimization at the second cycle provides a new optimal solution $\textbf{u}^2$, which contains optimum BHPs from day~$T_{cs}$ to day~$T_{N_t}$. These are applied until day~$2 \times T_{cs}$, after which the procedure is repeated until the last cycle of the CLRM is reached. In traditional CLRM (where high-fidelity simulation is applied for both production optimization and history matching), we generally observe that NPV increases over the prior optimum and uncertainty reduces as we proceed through the cycles. These trends are often not monotonic, however, as the models used at the different steps vary, and past well settings cannot be modified. We will observe similar behavior with proxy-based CLRM.

\section{Numerical Results}
\label{sec:4}
In this section, we first present results for CNN--RNN proxy well-rate predictions for multiple realizations for different BHP schedules. The proxy is then used, with retraining, for robust production optimization in a CLRM setting. 

\subsection{Model Setup}
\label{sec:4.1}
We consider 3D reservoir models characterized by Gaussian log-permeability fields. Each realization contains 40~$\times$~40~$\times$~8 grid blocks (12,800 total cells), with the blocks of dimensions 15~m~$\times$~15~m~$\times$~4~m. A total of 405 geological realizations of log-permeability were generated using sequential Gaussian simulation in SGeMS~\citep{SGeMs}. The realizations are conditioned to hard data at all locations penetrated by wells. A spherical variogram, with $r_{max}$ oriented 30$^{\circ}$ relative to the $y$-axis, and maximum, mid-range and minimum correlation lengths of $r_{max} = 25\Delta x$, $r_{mid} = 8\Delta x$, and $r_{min} = 2\Delta z$, was applied. The mean and standard deviation of log-permeability are 4.79 and 1.5, respectively. Porosity is constant at 0.2.

\begin{figure}[htbp!]\centering
\includegraphics[width=8cm]{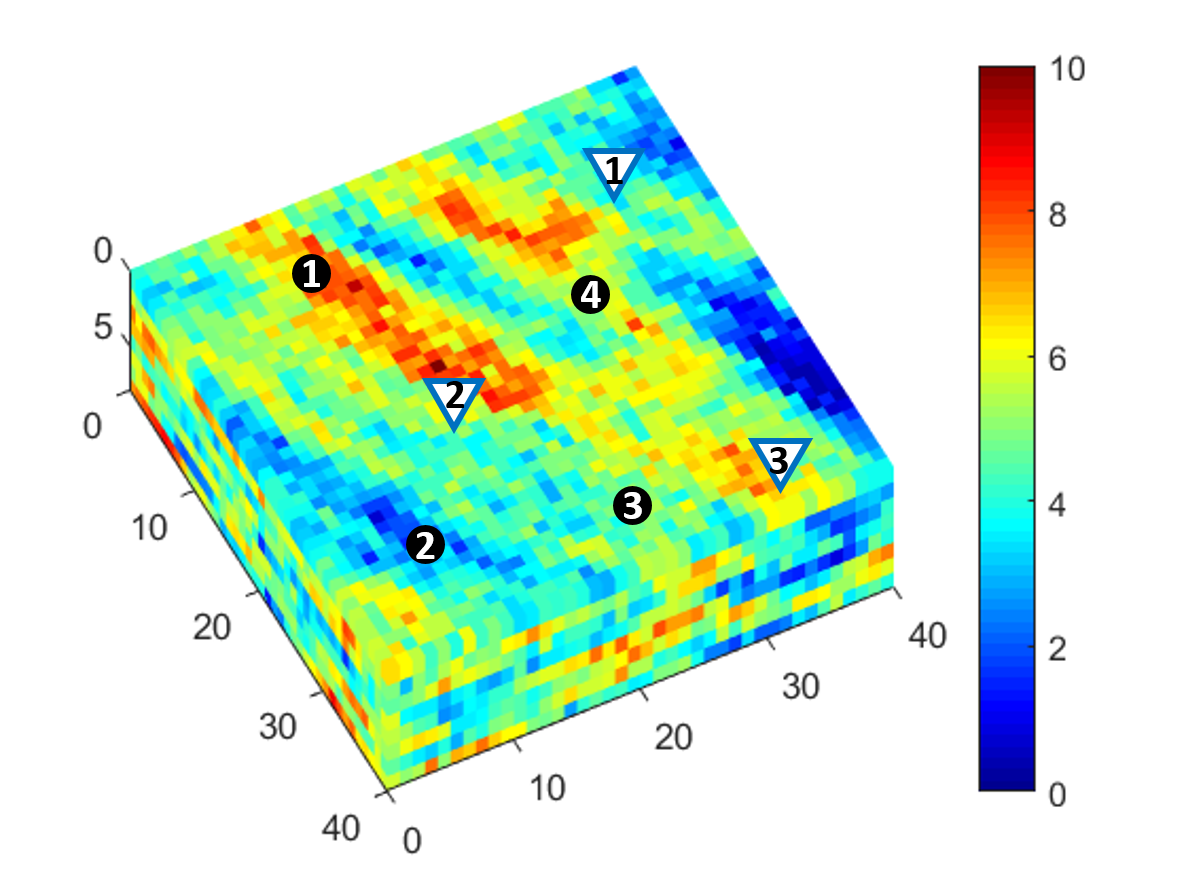}
\caption{Log-permeability field ($\log_e k$, with $k$ in md) for one realization. Permeability is conditioned to hard data at wells. Fully penetrating vertical production wells (black circles) and injection wells (white inverted triangles) are shown.} \label{fig:TrueModelA}
\end{figure}

A particular realization, which corresponds to one of the true models considered later, is shown in Figure~\ref{fig:TrueModelA}. The four black circles denote production wells, and the three white inverted triangles indicate injection wells. All wells are vertical, fully penetrating, and perforated over the entire reservoir thickness.

\begin{figure*}
\begin{subfigure}{.32\textwidth}
  \centering
  \includegraphics[width=1\linewidth]{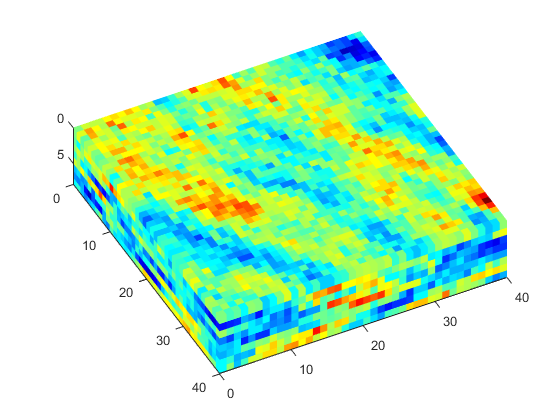}  
  \caption{Realization 1}
  \label{fig:ENSEM1}
\end{subfigure}
\begin{subfigure}{.32\textwidth}
  \centering
  \includegraphics[width=1\linewidth]{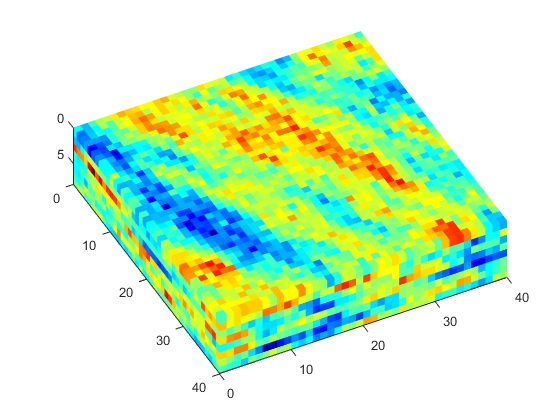}
  \caption{Realization 4}
  \label{fig:ENSEM4}
\end{subfigure}
\begin{subfigure}{.32\textwidth}
  \centering
  \includegraphics[width=1\linewidth]{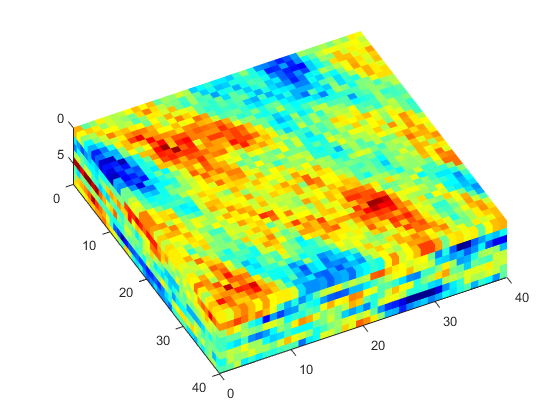}
  \caption{Realization 10}
  \label{fig:ENSEM10}
\end{subfigure}
\caption{Three prior log-permeability realizations ($\log_e k$, with $k$ in md). The colorbar in Figure~\ref{fig:TrueModelA} also applies here.}\label{fig:RealizationEnsem}
\end{figure*}

The first 400 of the 405 realizations are used to construct the PCA parameterization ($n_{rp}=400$). In this parameterization, we preserve 85\% of the `energy' computed from the $\Sigma$ matrix. This results in a reduction of the parameter space from 12,800 to 219. The remaining five realizations are used as true models in the CLRM examples. The first 20 realizations are also used as prior realizations for the initial (first CLRM cycle) robust production optimization and history matching procedures. Three of these prior realizations (without wells) are shown in Figure~\ref{fig:RealizationEnsem}. It is evident that, although all models honor hard data at the seven well locations, there is still significant variation between realizations.

\begin{figure}[htbp!]\centering
\includegraphics[width=8cm]{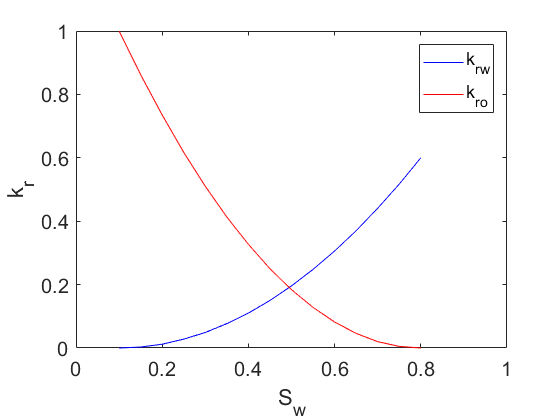}
\caption{Oil-water relative permeability curves.} \label{fig:twoD/RealtivePerm}
\end{figure}

In the simulations, the initial reservoir pressure is 320~bar and the initial oil saturation is 0.9. Oil and water viscosities are $\mu_o=2$~cp and $\mu_w=1$~cp. The oil (red) and water (blue) relative permeability curves are shown in Figure~\ref{fig:twoD/RealtivePerm}. Capillary pressure effects are neglected in this study.

All well are BHP controlled, with injector BHPs ranging from 325 to 335~bar and producer BHPs  from 300 to 315~bar. The BHPs are changed every 180~days, or a total of five times ($n_{cs}=5$), over the 900~day simulation time frame. The numbers of BHP optimization variables in each closed-loop cycle are 35, 28, 21, 14, and 7 (7~wells times the number of remaining control periods).

\subsection{Performance of CNN--RNN Proxy}

To train the proxy model, we randomly generate 500 different time-varying BHP profiles.
These profiles are then used with 20 different permeability realizations, with each realization assigned a different group of 25 BHP schedules (each schedule is used only once). We then perform a flow simulation for each geomodel--BHP profile set (for a total of 500 runs). From each simulation run, we extract oil and water rates every 30~days, from day~0 to day~900. Therefore, the number of total RNN time steps ($N_t$) is 31. For the four producers and three injectors, we thus have $N_{in}=7$ inputs and $N_{out}=11$ outputs at each time step. We set $N_{neu}=200$. The CNN reduces the 3D permeability information into two vectors, each of size of $N_{neu}$, which represent the initial long and short-term state vectors. We train the proxy using 300 simulation runs ($N_{B,train}=15$), while the remaining 200 runs ($N_{B,test}=10$) comprise the validation set. Training is performed to minimize $L$ (which is very similar to $E$ in Equation~\eqref{E}), and is terminated when $E\leq 0.05$. 

\begin{figure}[htbp!]\centering
\includegraphics[width=8cm]{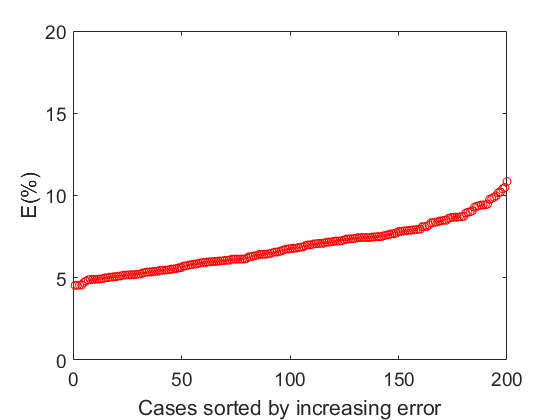}
\caption{Overall error $E^i$ for all test cases, sorted in increasing order. These results involve 20 different permeability realizations and 200 different BHP profiles.} \label{fig:error}
\end{figure}

We now assess the performance of the trained proxy. Figure~\ref{fig:error} shows the overall time-average error $E^i$, given by Equation~\eqref{Ei}, for all 200 test cases. These error values are sorted in increasing order. The 10th, 50th and 90th percentile errors are 5.1\%, 6.8\% and 8.7\%. These values, and the results in Figure~\ref{fig:error}, indicate that the proxy performs reasonably well over a range of realizations and BHP profiles. We obtained similar errors for the single realization case considered in \citep{Kim2021RNN}. Specifically, for a 2D Gaussian case assessed in that study, the 10th, 50th and 90th percentile errors were 5.2\%, 6.1\% and 7.4\%. 

\begin{figure*}[htbp!]\centering
\begin{subfigure}{6.5cm}
\includegraphics[width=1\textwidth]{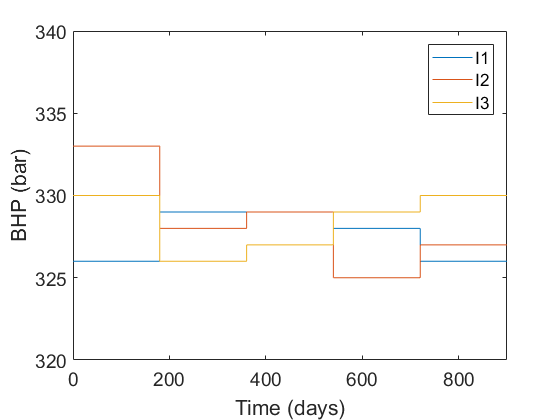}
\caption{INJ BHPs}
\label{fig:p50INJ}
\end{subfigure}
\begin{subfigure}{6.5cm}
\includegraphics[width=1\textwidth]{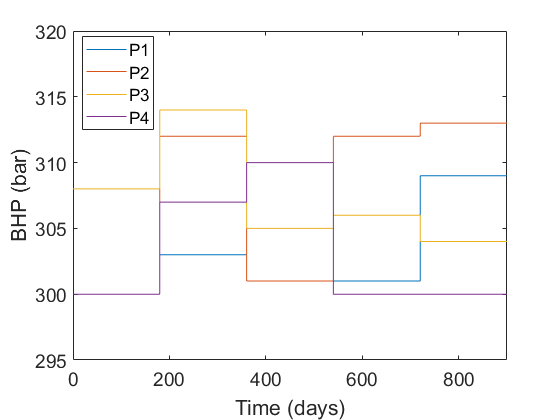}
\caption{PRD BHPs}
\label{fig:p50PRD}
\end{subfigure}
\begin{subfigure}{6.5cm}
\includegraphics[width=1\textwidth]{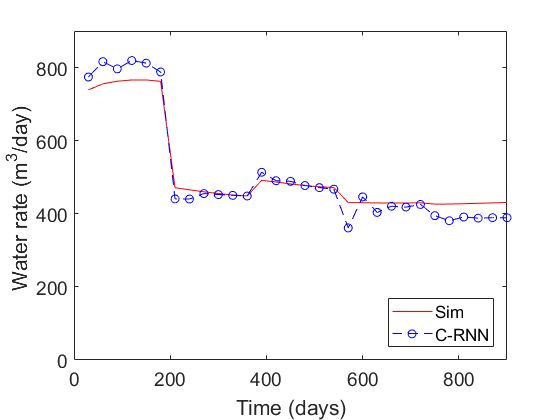}
\caption{Water rate (INJ2)}
\label{fig:p50k}
\end{subfigure}
\begin{subfigure}{6.5cm}
\includegraphics[width=1\textwidth]{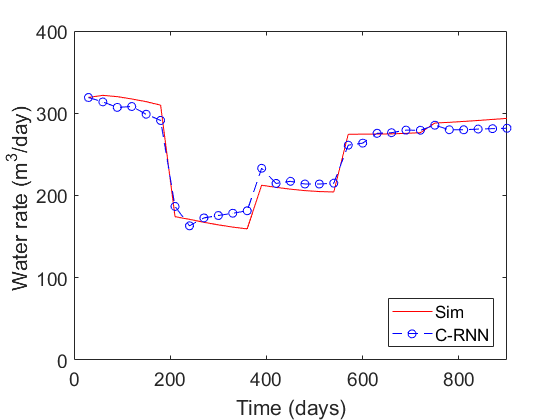}
\caption{Water rate (INJ3)}
\label{fig:p50I2water}
\end{subfigure}
\begin{subfigure}{6.5cm}
\includegraphics[width=1\textwidth]{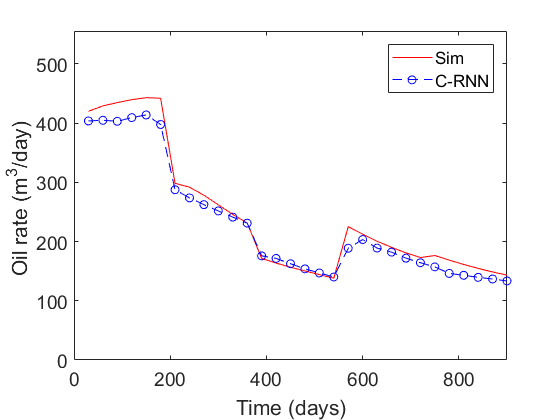}
\caption{Oil rate (PRD4)}
\label{fig:p50P4oil}
\end{subfigure}
\begin{subfigure}{6.5cm}
\includegraphics[width=1\textwidth]{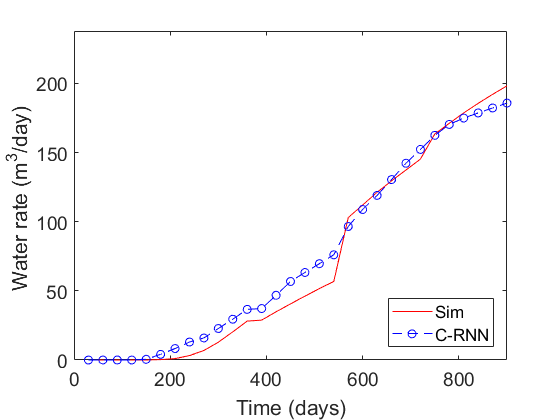}
\caption{Water rate (PRD4)}
\label{fig:p50P4water}
\end{subfigure}
\begin{subfigure}{6.5cm}
\includegraphics[width=1\textwidth]{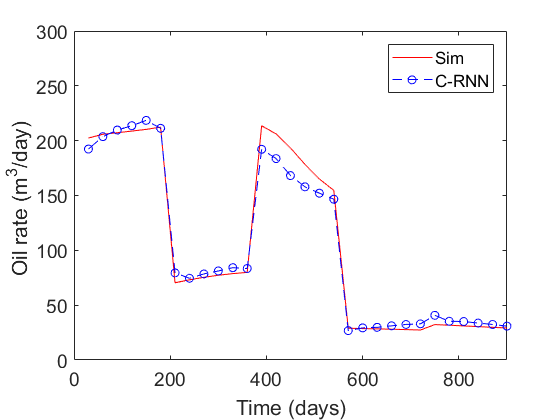}
\caption{Oil rate (PRD2)}
\label{fig:p50P2oil}
\end{subfigure}
\begin{subfigure}{6.5cm}
\includegraphics[width=1\textwidth]{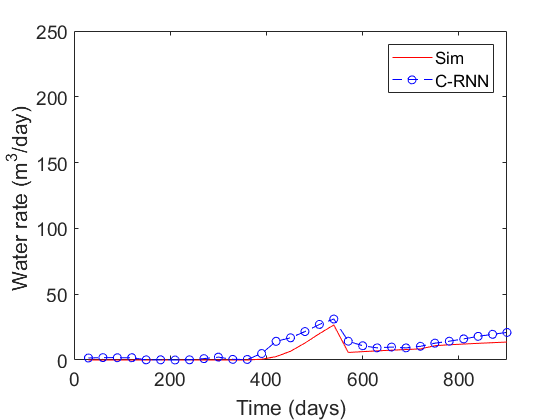}
\caption{Water rate (PRD2)}
\label{fig:p50P2water}
\end{subfigure}
\caption{Input BHP profiles and corresponding well rates from high-fidelity reservoir simulation (red curves) and CNN-RNN proxy (blue curves with circles) for the 50th percentile error test case. Permeability field for this case shown in Figure~\ref{fig:ENSEM4}. INJ2 and INJ3 are the highest cumulative water injection wells, PRD4 is the highest cumulative oil production well, and PRD2 is the lowest cumulative oil production well.} 
\label{fig:P50 test case}
\end{figure*}

\begin{figure*}[htbp!]\centering
\begin{subfigure}{6.5cm}
\includegraphics[width=1\textwidth]{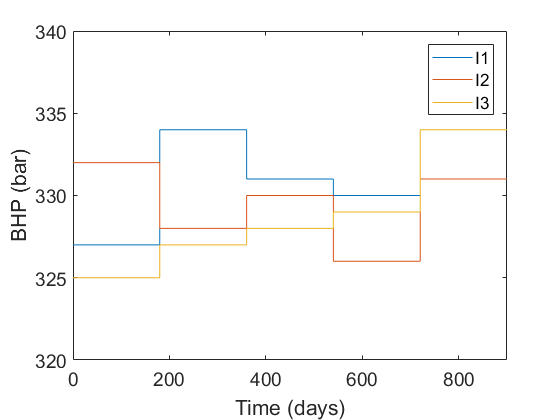}
\caption{INJ BHPs}
\label{fig:p60INJ}
\end{subfigure}
\begin{subfigure}{6.5cm}
\includegraphics[width=1\textwidth]{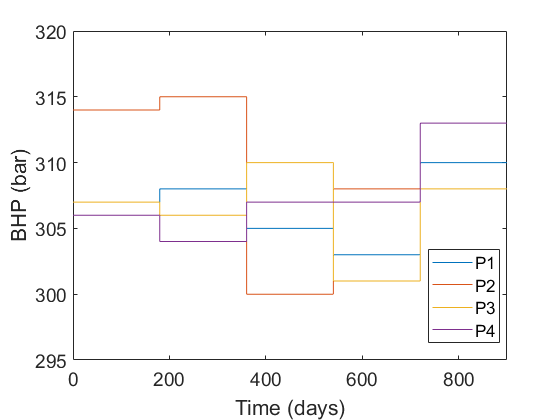}
\caption{PRD BHPs}
\label{fig:p60PRD}
\end{subfigure}
\begin{subfigure}{6.5cm}
\includegraphics[width=1\textwidth]{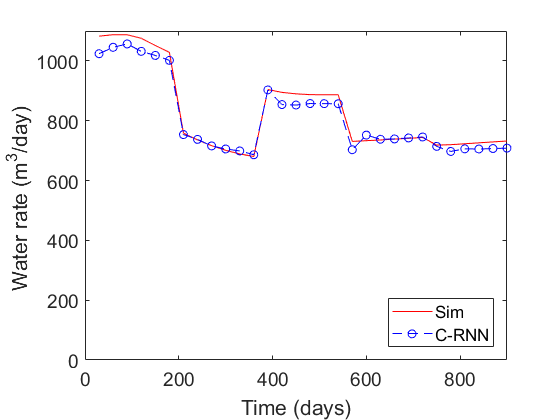}
\caption{Water rate (INJ2)}
\label{fig:p60I2}
\end{subfigure}
\begin{subfigure}{6.5cm}
\includegraphics[width=1\textwidth]{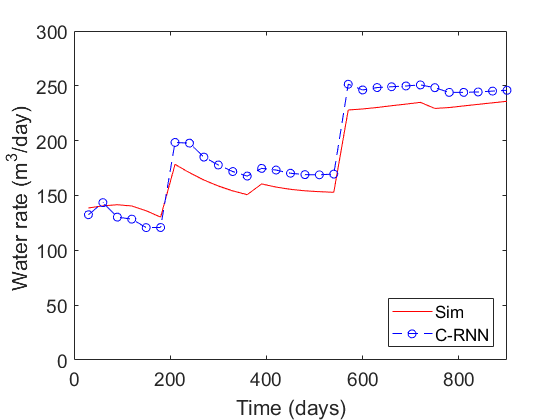}
\caption{Water rate (INJ3)}
\label{fig:p60I3}
\end{subfigure}
\begin{subfigure}{6.5cm}
\includegraphics[width=1\textwidth]{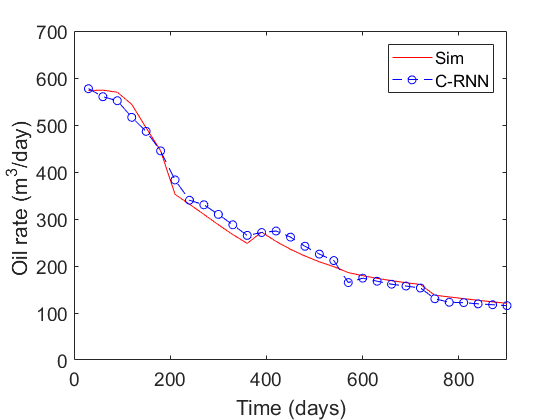}
\caption{Oil rate (PRD1)}
\label{fig:p60P1oil}
\end{subfigure}
\begin{subfigure}{6.5cm}
\includegraphics[width=1\textwidth]{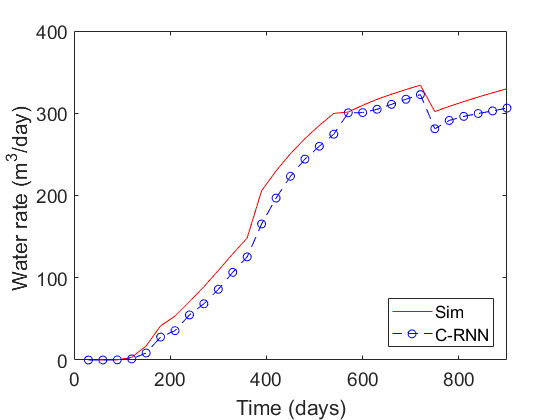}
\caption{Water rate (PRD1)}
\label{fig:p60P1water}
\end{subfigure}
\begin{subfigure}{6.5cm}
\includegraphics[width=1\textwidth]{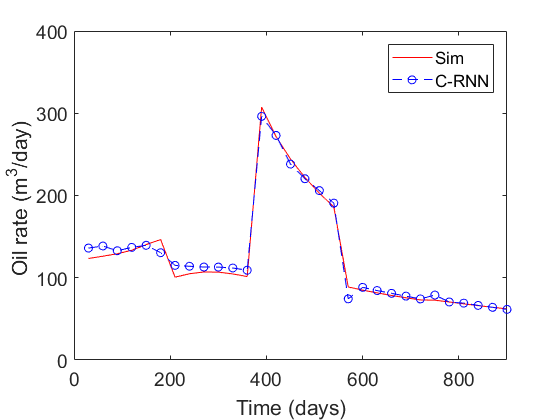}
\caption{Oil rate (PRD2)}
\label{fig:p60P2oil}
\end{subfigure}
\begin{subfigure}{6.5cm}
\includegraphics[width=1\textwidth]{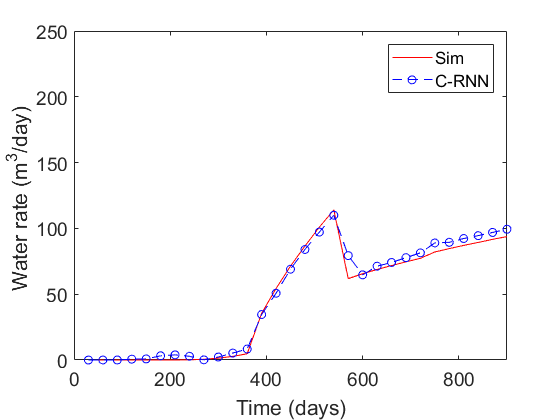}
\caption{Water rate (PRD2)}
\label{fig:p60P2water}
\end{subfigure}
\caption{Input BHP profiles and corresponding well rates from high-fidelity reservoir simulation (red curves) and CNN-RNN proxy (blue curves with circles) for the 60th percentile error test case. Permeability field for this case shown in Figure~\ref{fig:ENSEM10}. INJ2 and INJ3 are the highest cumulative water injection wells, PRD1 is the highest cumulative oil production well, and PRD2 is the lowest cumulative oil production well.} 
\label{fig:P60 test case}
\end{figure*}

We now show results for cases with errors $E^i$ near the median. Figures~\ref{fig:P50 test case} and \ref{fig:P60 test case} show input BHP profiles and output well rates for the 50th and 60th percentile error cases (the latter corresponds to 7.2\% error). Permeability fields for these cases appear in Figure~\ref{fig:ENSEM4} and \ref{fig:ENSEM10}. In the well-rate plots, the red curves indicate the reference numerical simulation results and the blue curves with the circles depict the CNN--RNN proxy results. Results for the two injectors with the highest cumulative water injection are shown, along with oil and water rates for the producers with the highest and lowest cumulative oil production. The results in both figures demonstrate that the CNN--RNN proxy is able to predict well-by-well oil and water rates, including water breakthrough times, for different BHP profiles and different realizations. Results are presented in both figures for injectors INJ2 and INJ3 and for producer PRD2. The rate profiles for these wells are very different between the two figures, demonstrating the wide range of behavior the proxy model is able to capture.

\begin{figure*}
\begin{minipage}{.33\textwidth}
  \centering
  \includegraphics[width=1\linewidth]{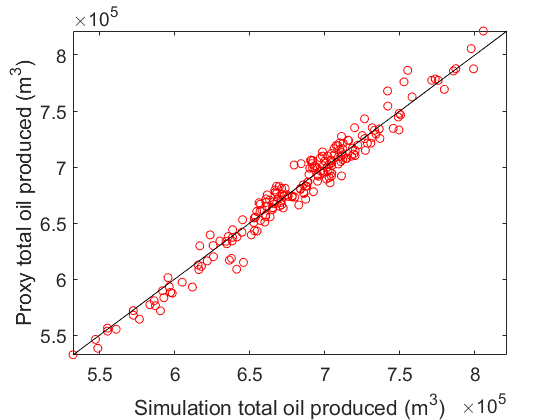}  
  \subcaption{Cumulative field-wide oil produced}
\end{minipage}
\begin{minipage}{.33\textwidth}
  \centering
  \includegraphics[width=1\linewidth]{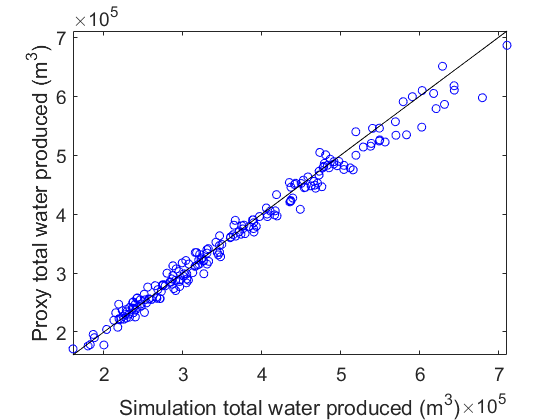}
  \subcaption{Cumulative field-wide water produced}
\end{minipage}
\begin{minipage}{.33\textwidth}
  \centering
  \includegraphics[width=1\linewidth]{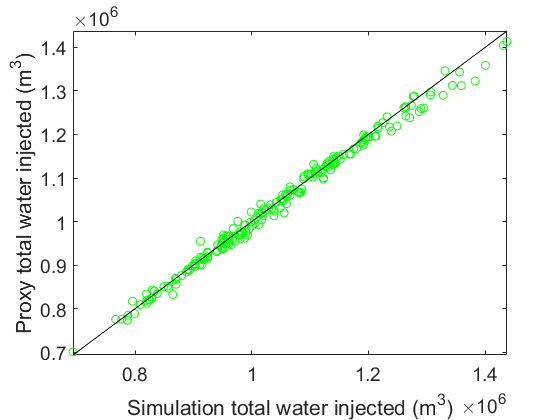}
  \subcaption{Cumulative field-wide water injected}
\end{minipage}
\caption{Test-case cross-plots for cumulative field-wide oil and water production and water injection.}
\label{fig:Cum}
\end{figure*}

Additional comparisons, for all 200 test cases, are shown in Figure~\ref{fig:Cum}. There we present cross-plots (proxy model results versus simulation results) for cumulative field-wide oil and water produced and water injected. These cumulative field-wide quantities are calculated from the well-by-well time-series results. The points in the three plots generally fall near the 45$^{\circ}$ line, again demonstrating the effectiveness of the proxy model. Some deviation is evident in the water production results (Figure~\ref{fig:Cum}b), though it is important to note that these results span a large range -- about a factor of four -- in cumulative water produced.

\subsection{CLRM Results using CNN--RNN Proxy}
In this section we apply the CNN--RNN proxy in a CLRM workflow. The CLRM is run five separate times, each time with a different (synthetic) `true' model. The true models, used to provide the observation data, are referred to as True model~A through True model~E. The same 20 initial prior models are used in all cases. For the NPV calculation, the price of oil is set to \$74/STB, the costs of injected and produced water are \$9/STB and \$5/STB, respectively, and the discount rate is 0.1.

The nonlinear output constraints applied in this study are a maximum field water injection rate of 1400~m$^3$/day, a maximum well water injection rate of 1100~m$^3$/day, and a maximum field water production rate of 1100~m$^3$/day. Thus, we have a total of five nonlinear output constraints. During robust production optimization, these constraints are handled using the filter-based procedure within PSO, as described in Section~\ref{sec:3.2}. For each CLRM cycle, we run PSO for $N_i=30$ iterations with a swarm size of $N_s=35$.

\begin{figure*}
\begin{minipage}{.33\textwidth}
  \centering
  \includegraphics[width=1\linewidth]{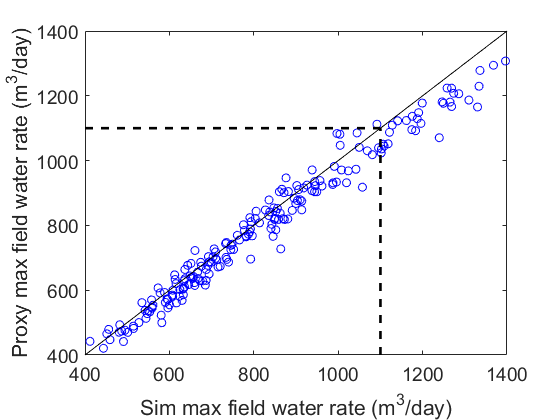}  
  \subcaption{Maximum field water production rate}
\end{minipage}
\begin{minipage}{.33\textwidth}
  \centering
  \includegraphics[width=1\linewidth]{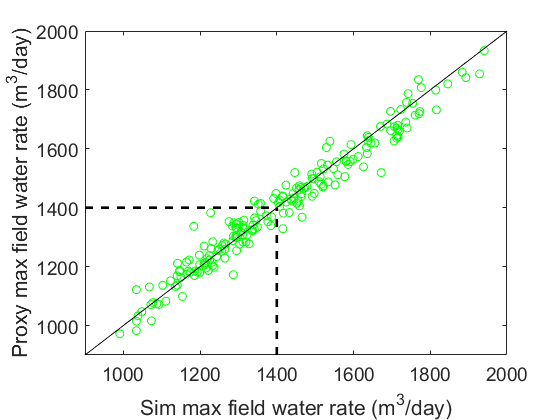}
  \subcaption{Maximum field water injection rate}
\end{minipage}
\begin{minipage}{.33\textwidth}
  \centering
  \includegraphics[width=1\linewidth]{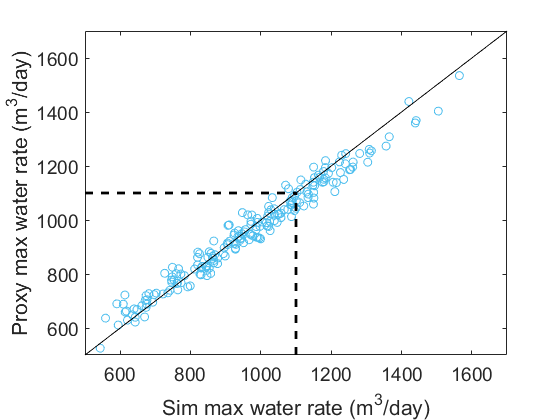}
  \subcaption{Maximum injection rate for INJ2}
\end{minipage}
\caption{Test-case cross-plots for maximum field water production, field water injection, and INJ2 rates. Results in all cases are the maximum value observed over the full simulation time frame. Dashed lines indicate nonlinear constraint limits applied in CLRM runs.}
\label{fig:constraintCros}
\end{figure*}

Before presenting CLRM results, we demonstrate the ability of the proxy model to provide accurate estimates of the quantities required for nonlinear constraint evaluation. Figure~\ref{fig:constraintCros} displays cross-plots of three such quantities -- maximum field water production rate, maximum field water injection rate, and INJ2 maximum injection rate -- for the 200 test cases. These results correspond to the maximum observed rate for each quantity over the full simulation time frame. The dashed lines indicate the limits specified by the nonlinear output constraints. These cross-plots demonstrate that the proxy provides reasonably accurate estimates of these rates, both at the well and field level, indicating that it can indeed give quantitative constraint violation information. It is important to note that well settings that lead to violations of the nonlinear constraints are observed in many cases, suggesting that these constraints will be active during the course of the CLRM optimizations. 

For history matching, the well-by-well oil and water injection and production rates are recorded, from the simulation of the true model, every 90~days. The observed data are obtained by adding random (Gaussian) noise to the true-model simulation results. The standard deviation of the Gaussian noise is set to 2\% of the simulated value. We run the RML history matching procedure $n_r=20$ times to generate 20 posterior realizations. At each 90-day interval, three water injection rates, four oil production rates, and four water production rates are observed. Thus the number of observations used for history matching at the various CLRM cycles are 22 (at day~180), 44 (at day~360), 66 (at day~540), and 88 (at day~720).

\begin{figure*}[htbp!]\centering
\begin{subfigure}{12.5cm}
\includegraphics[width=1\textwidth]{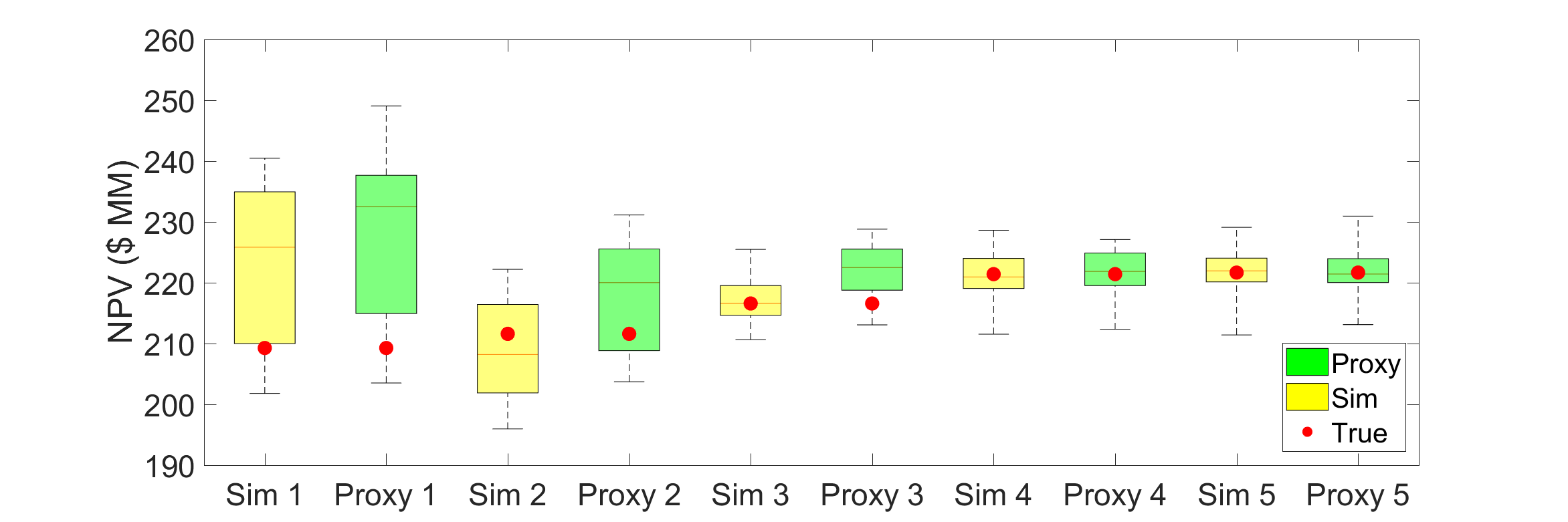}
\caption{True model A}
\label{fig:CLRM_TA}
\end{subfigure}
\begin{subfigure}{12.5cm}
\includegraphics[width=1\textwidth]{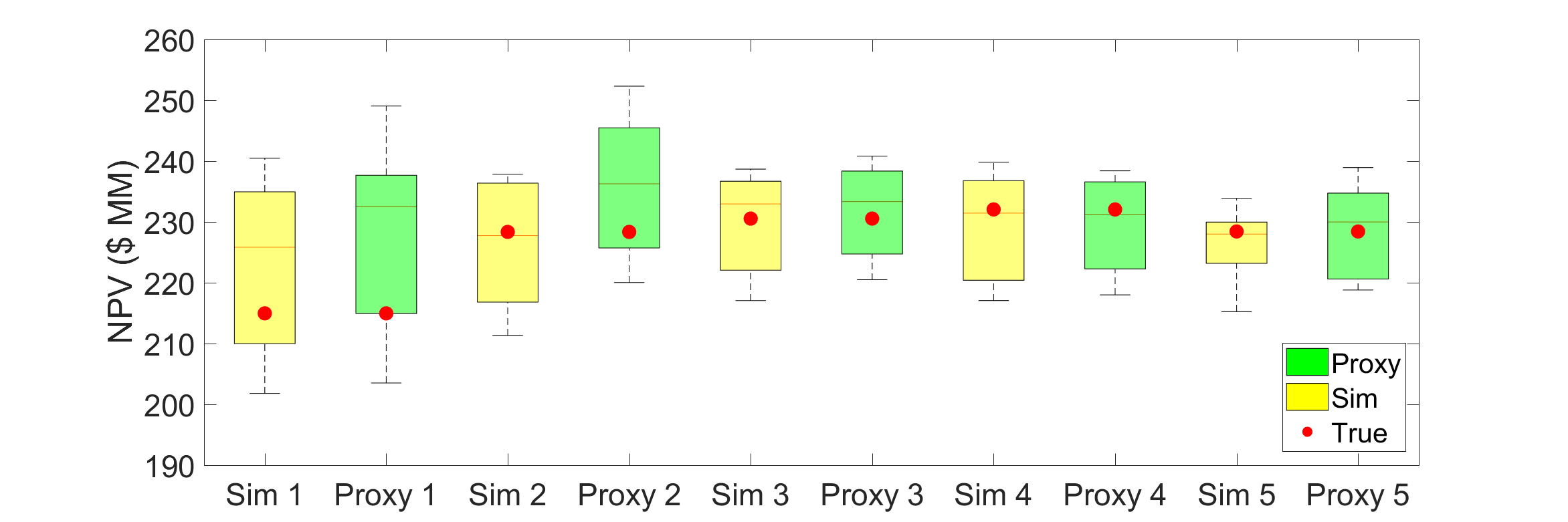}
\caption{True model B}
\label{fig:CLRM_TB}
\end{subfigure}
\begin{subfigure}{12.5cm}
\includegraphics[width=1\textwidth]{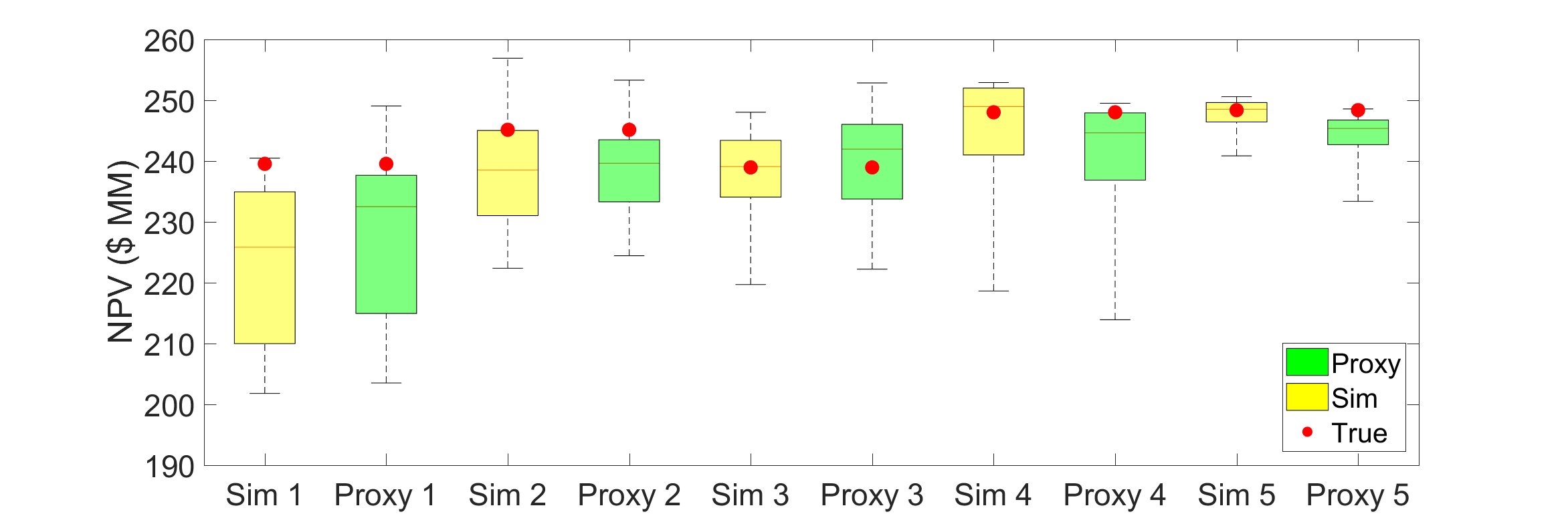}
\caption{True model C}
\label{fig:CLRM_TC}
\end{subfigure}
\begin{subfigure}{12.5cm}
\includegraphics[width=1\textwidth]{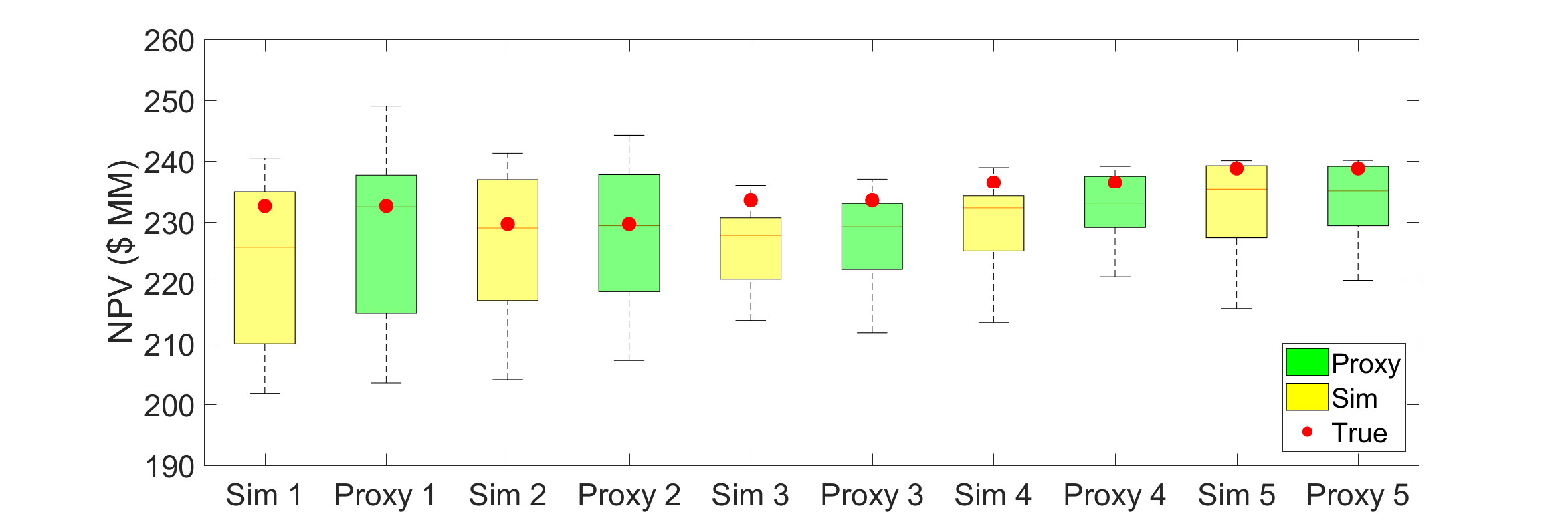}
\caption{True model D}
\label{fig:CLRM_TD}
\end{subfigure}
\begin{subfigure}{12.5cm}
\includegraphics[width=1\textwidth]{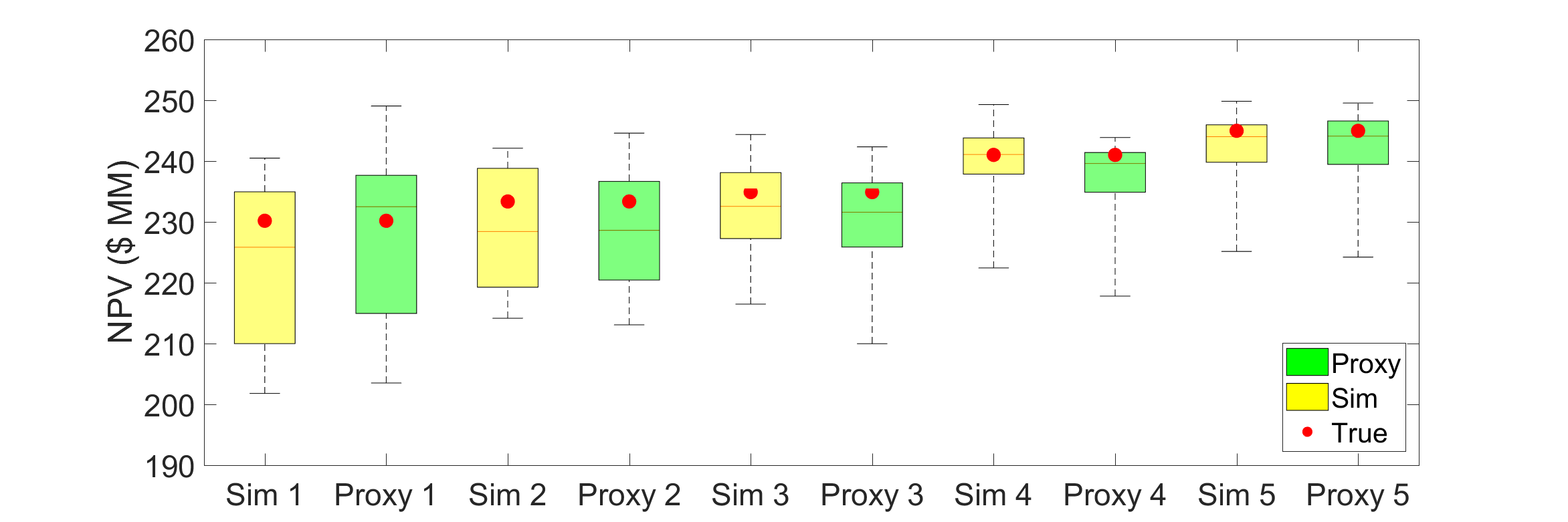}
\caption{True model E}
\label{fig:CLRM_TE}
\end{subfigure}
\caption{Box plots of NPVs for True models~A–E at different CLRM cycles (see text for description of percentile quantities displayed in boxes). Green boxes show results for the CNN--RNN proxy and yellow boxes are simulation results using the $n_{ro} = 16$ current models and BHPs from proxy-based optimization. Red circles indicate numerical simulation results for the true model.} 
\label{fig:boxplots}
\end{figure*}

Figure~\ref{fig:boxplots} displays box plots of NPVs for the optimized BHPs at different CLRM cycles, for True models~A-E. The red lines inside the boxes indicate the median NPV from the $n_r=20$ realizations, and the bottom and top of the boxes represent the 25th and 75th percentile results over the 20 models. The lines extending outside the boxes depict the 10th and 90th percentile results of the $n_r=20$ realizations. Thus, Figure~\ref{fig:boxplots} displays the full range of NPVs from the $n_{ro}=16$ realizations used for optimization and constraint satisfaction. The green boxes represent results from the CNN--RNN proxy, and the yellow boxes display simulation results obtained by simulating the $n_{ro} = 16$ current models using the optimized BHPs determined from proxy-based optimization. The red circles show the simulated NPV for the particular true model at the current CLRM cycle (at each cycle, these are the same in the yellow and green boxes).

For the five true models, the green boxes are all the same at CLRM cycle~1 because the first robust optimization is performed using the (same) $n_{ro}$ realizations. The yellow boxes are also the same at cycle~1 for all true models because the same controls (based on prior model optimization) are used for each model. Although the controls are the same, the NPVs for True models~A-E (red circles) differ from one another at cycle~1 because the models are different. The first history matching is performed at day~180 (start of CLRM cycle~2), after we operate the field based on the optimum BHPs from the first CLRM cycle. Nonlinear constraints are enforced for the $n_{ro}$ realizations in every CLRM cycle.

From Figure~\ref{fig:boxplots}, we see that the proxy and simulation-based box plots of NPV generally coincide (though there are differences) and show similar ranges of uncertainty. As demonstrated earlier, the proxy provides reasonable predictions of oil and water rates over multiple realizations. Thus it is not surprising that it gives satisfactory estimates of the corresponding NPVs. We also observe in Figure~\ref{fig:boxplots} that the NPVs for the five true models generally trend upward as we proceed from cycle to cycle. The improvement in NPV, from CLRM cycle~1 to cycle~5 for the five  true models, is quantified in Table~\ref{NPVimprovement}. There we see improvements ranging from 2.7\% to 6.4\% from cycles~1 to 5.

As is typical in CLRM, despite achieving overall improvement in NPV, we do not always observe a monotonic increase in NPV. For example, there is a slight decrease in NPV for True model~C from cycle~2 to cycle~3 (Figure~\ref{fig:boxplots}c), though NPV for this case does increase at the next cycle. This non-monotonic behavior occurs because the models used for optimization change from cycle to cycle, and there is no guarantee that the updated models provide higher NPVs than the previous set of models. In addition, at later stages the remaining simulation time frame and the number of optimization variables decrease, so there is less scope for improvement. Uncertainty reduction is clearly observed (i.e., the sizes of the boxes decrease) as we proceed through the cycles, though again this decrease is not monotonic.

\begin{table}[h]
  \centering
  \caption{NPVs for the five true models at CLRM cycles~1 and 5.}
  \label{NPVimprovement}
  \begin{tabular}{c c c c c c }
  	\hline 
    True &  NPV at cyc 1  & NPV at cyc 5 & Improvement  \\
    model &  (10$^6$~USD) & (10$^6$~USD) & (\%) \\
     \hline
     \hline
      A &  209.3 & 221.7 & 5.92 \\
      B &  215.0 & 228.4 & 6.23 \\
      C &  239.5 & 248.4 & 3.72 \\
      D &  232.6 & 238.8 & 2.67 \\
      E &  230.2 & 245.0 & 6.43\\
     \hline
  \end{tabular}
\end{table}

\begin{figure*}[htbp!]\centering
\begin{subfigure}{8.6cm}
\includegraphics[width=\textwidth]{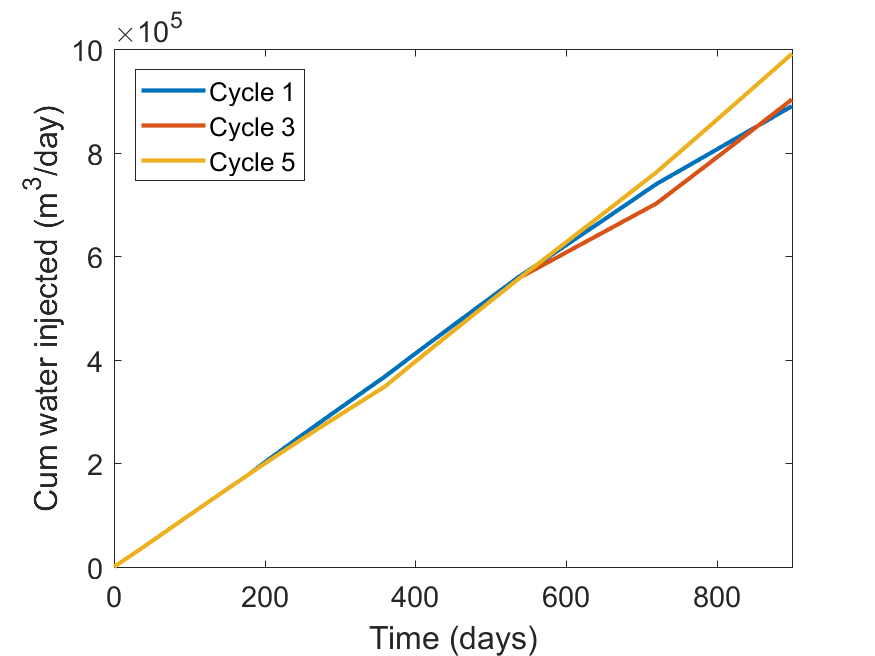}
\caption{Cumulative water injected} 
\label{fig:stepCWIMedian}
\end{subfigure}
\begin{subfigure}{8.6cm}
\includegraphics[width=\textwidth]{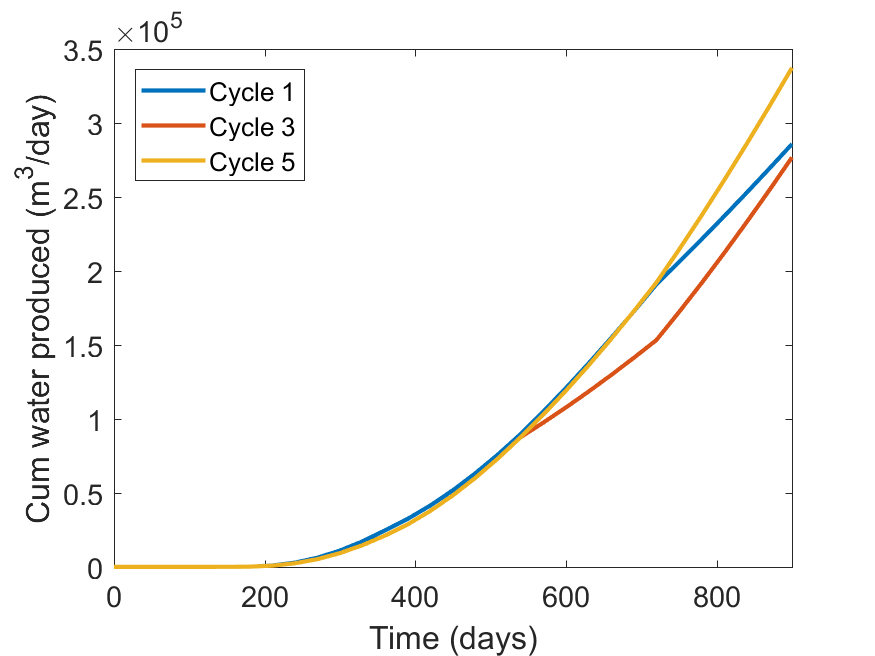}
\caption{Cumulative water produced} 
\label{fig:stepCWPMedian}
\end{subfigure}
\begin{subfigure}{8.6cm}
\includegraphics[width=\textwidth]{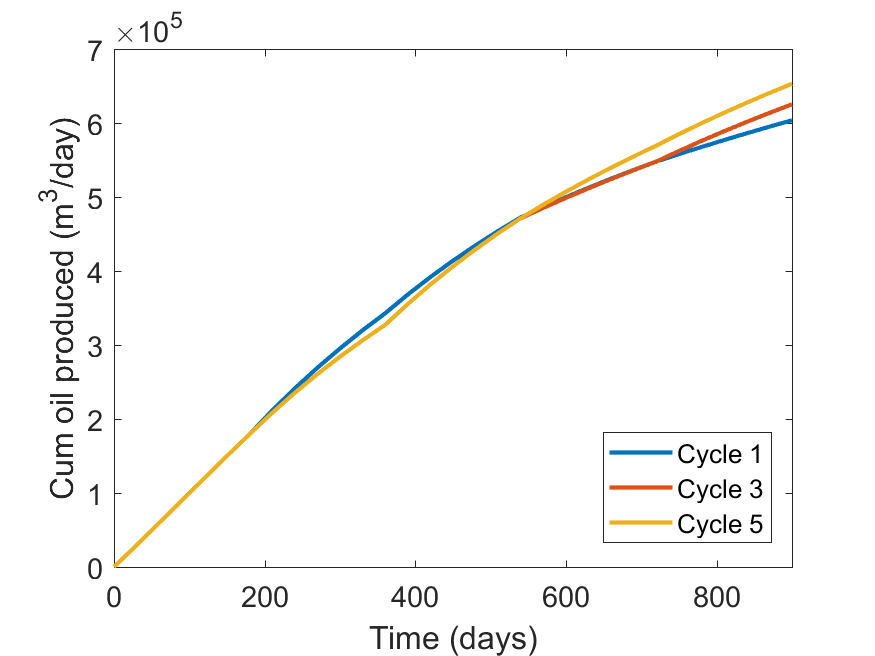}
\caption{Cumulative oil produced}
\label{fig:stepCOPMedian}
\end{subfigure}
\begin{subfigure}{8.6cm}
\includegraphics[width=\textwidth]{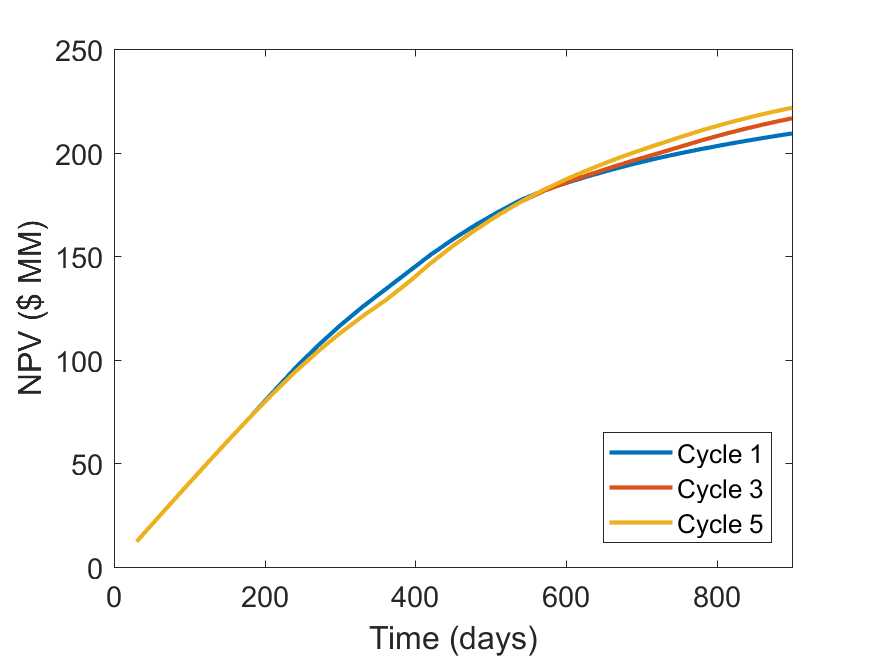}
\caption{NPV}
\label{fig:stepNPVMedian}
\end{subfigure}
\caption{Optimum solutions found by CNN-RNN proxy at CLRM cycles~1, 3 and 5 for True model~A. The results shown are obtained by performing simulation using BHPs from proxy-based optimization at different CLRM cycles.} \label{fig:stepcompareMedian}
\end{figure*}

Field-wide responses for True model~A are shown in Figure~\ref{fig:stepcompareMedian}. Note that True model~A corresponds to the median case in terms of improvement in NPV from CLRM cycle~1 to 5 (see Table~\ref{NPVimprovement}). Figure~\ref{fig:stepcompareMedian} displays the evolution of cumulative water injected, cumulative water produced, cumulative oil produced, and NPV, for CLRM cycles~1, 3 and 5. These results are obtained by simulating True model~A using the BHPs from proxy-based optimization. The increase in NPV is achieved by increasing cumulative oil production by 8.2\%. This is driven by an increase in cumulative water injection of 11.4\%, and it results in an 18.0\% increase in cumulative water production (from CLRM cycle~1 to 5).

\begin{figure*}[htbp!]\centering
\begin{subfigure}{8.6cm}
\includegraphics[width=\textwidth]{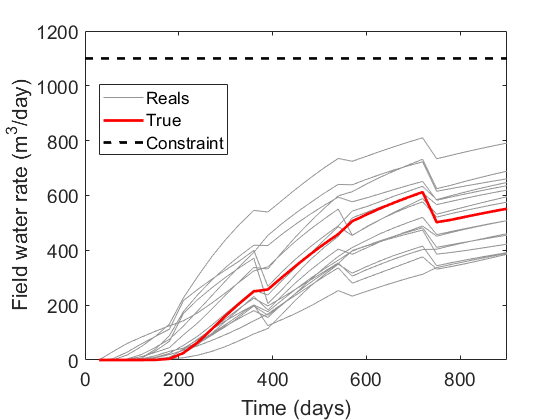}
\caption{Field water production rate at cycle~1} 
\end{subfigure}
\begin{subfigure}{8.6cm}
\includegraphics[width=\textwidth]{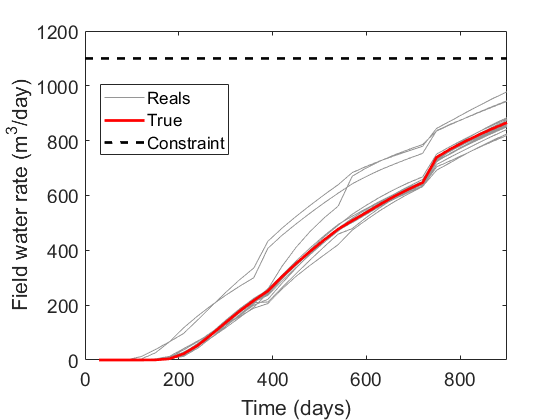}
\caption{Field water production rate at cycle~5} 
\end{subfigure}
\begin{subfigure}{8.6cm}
\includegraphics[width=\textwidth]{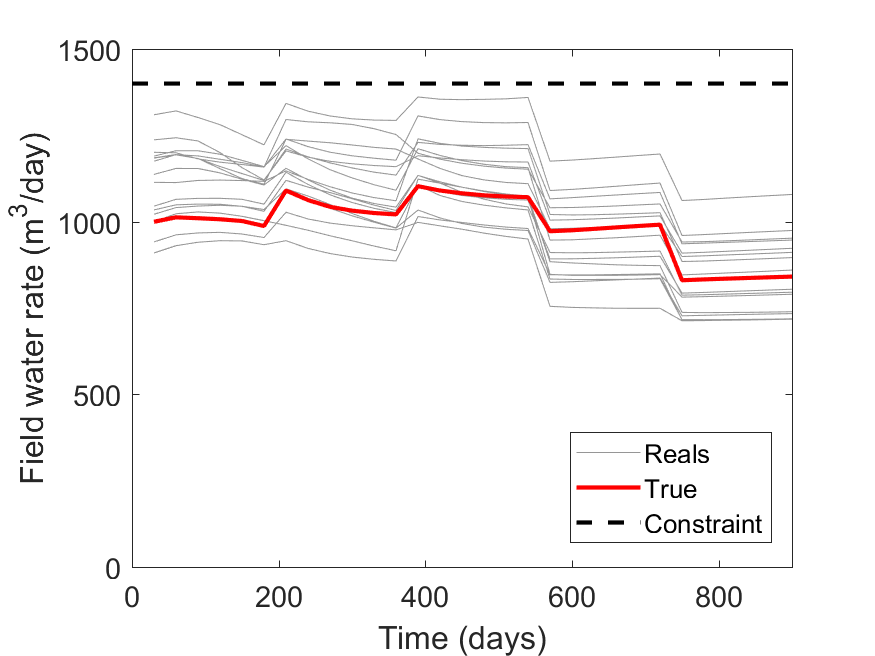}
\caption{Field water injection rate at cycle~1}
\end{subfigure}
\begin{subfigure}{8.6cm}
\includegraphics[width=\textwidth]{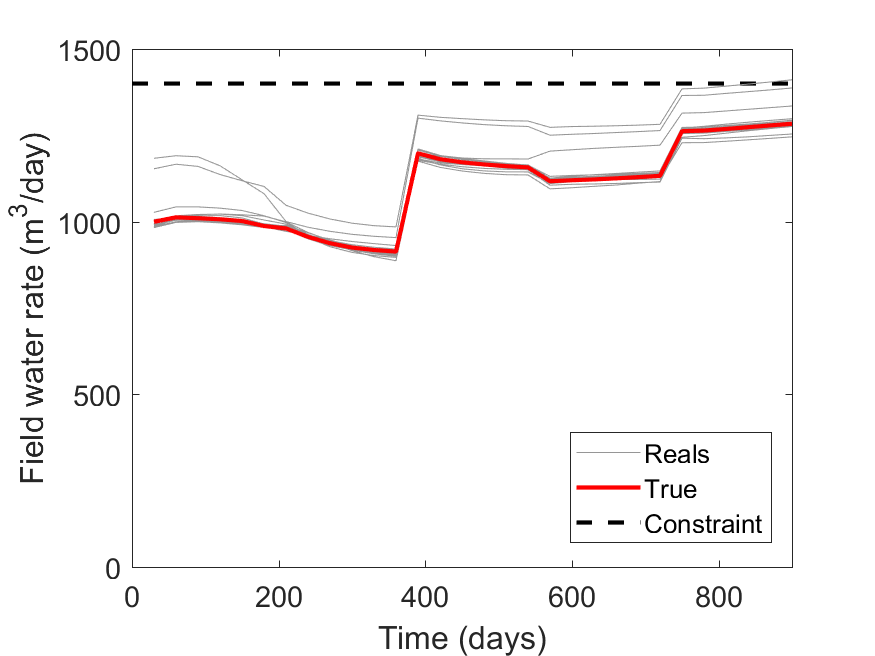}
\caption{Field water injection rate at cycle~5}
\end{subfigure}
\begin{subfigure}{8.6cm}
\includegraphics[width=\textwidth]{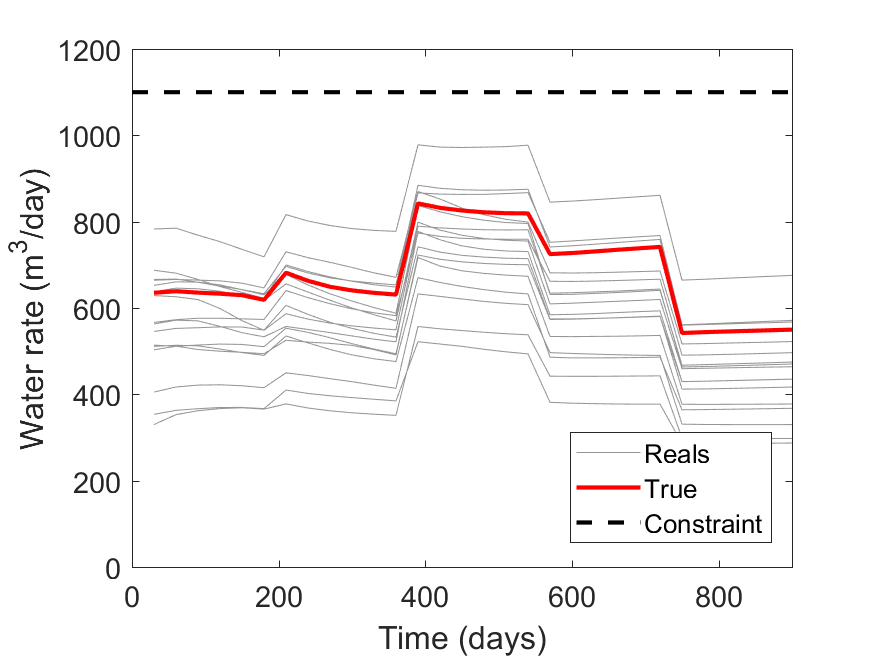}
\caption{INJ2 water rate at cycle 1}
\end{subfigure}
\begin{subfigure}{8.6cm}
\includegraphics[width=\textwidth]{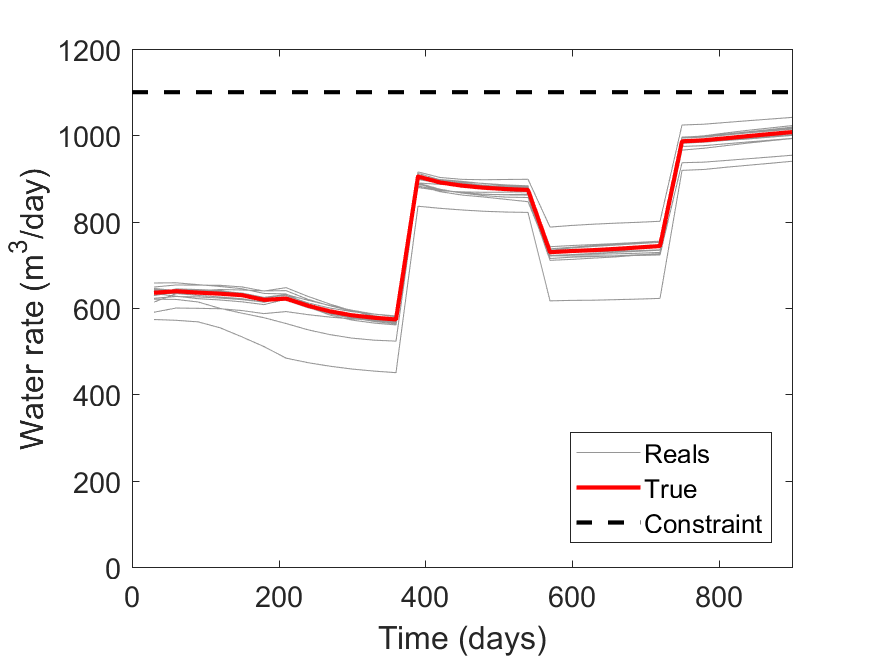}
\caption{INJ2 water rate at cycle 5}
\end{subfigure}
\caption{Optimized solutions at CLRM cycles~1 and 5 for True model~A. Results shown for $n_{ro}=16$ realizations (gray curves) and for true model. Nonlinear constraint limits shown as dashed lines.} \label{fig:constraintTrueA}
\end{figure*}

The range of solutions for the $n_{ro}$ realizations, and their feasibility in terms of the nonlinear constraints, is shown in Figure~\ref{fig:constraintTrueA}. Here we show field-wide water production rate, field-wide water injection rate, and INJ2 water rate for CLRM cycle~1 (left plots) and cycle~5 (right plots). The gray lines represent solutions from the $n_{ro}$ realizations considered in the robust optimization, the red curve shows the response of True model~A, and the dashed lines indicate the constraint limits. Consistent with the results in Figure~\ref{fig:stepcompareMedian}, these results are obtained from high-fidelity simulation of True model~A and the $n_{ro}$ realizations using the BHPs from proxy-based optimization. We see that the constraints are satisfied for the $n_{ro}$ realizations and for True model~A in all cases. It is not assured, however, that the constraints will be satisfied for the true model even if they are satisfied for all $n_{ro}$ realizations, and we have observed small constraint violations at some control steps for some true models. 

The reduction in variation over the $n_{ro}$ realizations, from cycle~1 to cycle~5, is apparent for all three constraint quantities. We also see that the solution at cycle~1 must be `conservative' to avoid violating constraints over all models. By cycle~5, however, there is less uncertainty in the geological model, and the optimization can `push' the solutions close to the constraint limit.

\begin{figure*}[htbp!]\centering
\begin{subfigure}{8.6cm}
\includegraphics[width=\textwidth]{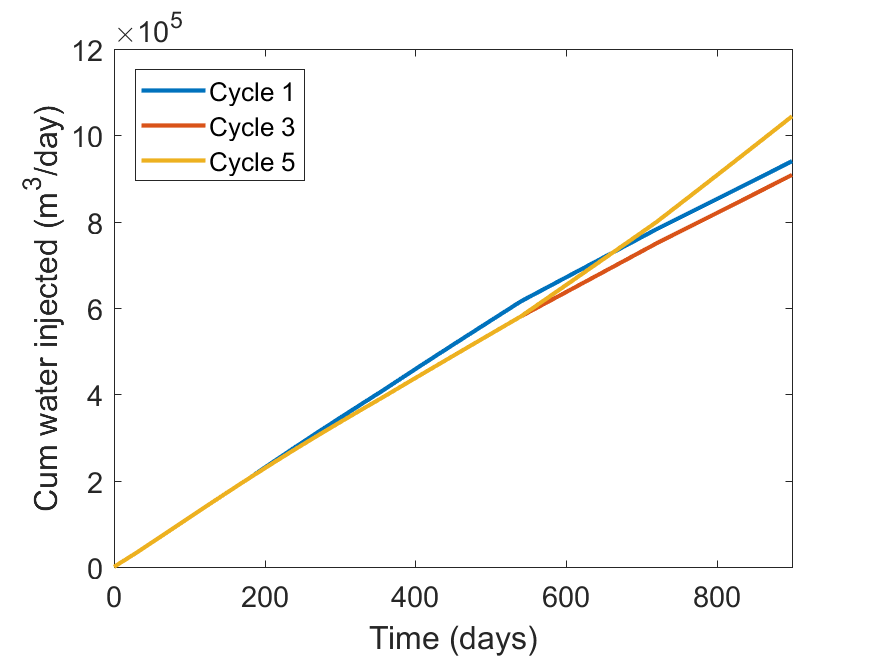}
\caption{Cumulative water injected} 
\label{fig:stepCWI_C}
\end{subfigure}
\begin{subfigure}{8.6cm}
\includegraphics[width=\textwidth]{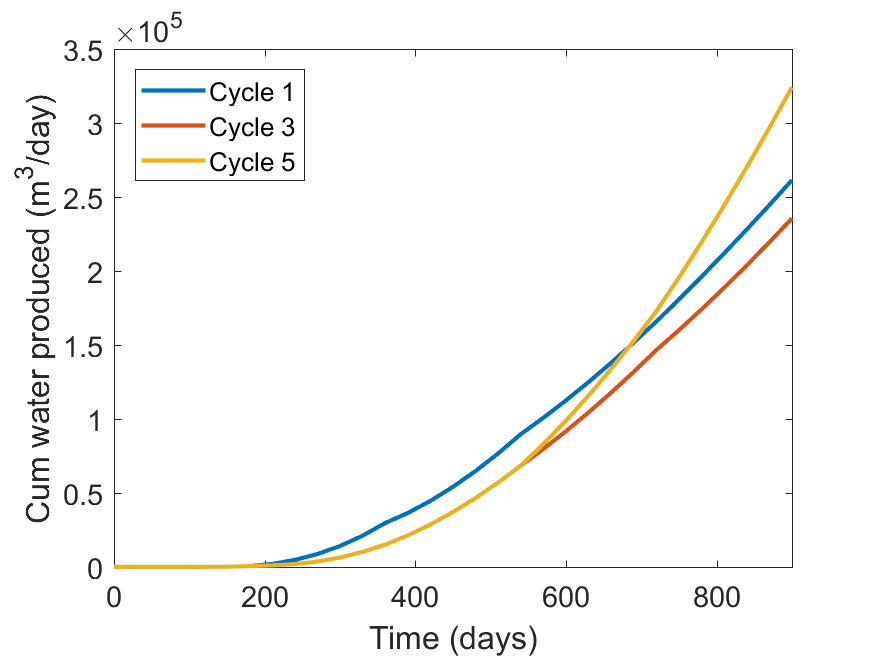}
\caption{Cumulative water produced} 
\label{fig:stepCWP_C}
\end{subfigure}
\begin{subfigure}{8.6cm}
\includegraphics[width=\textwidth]{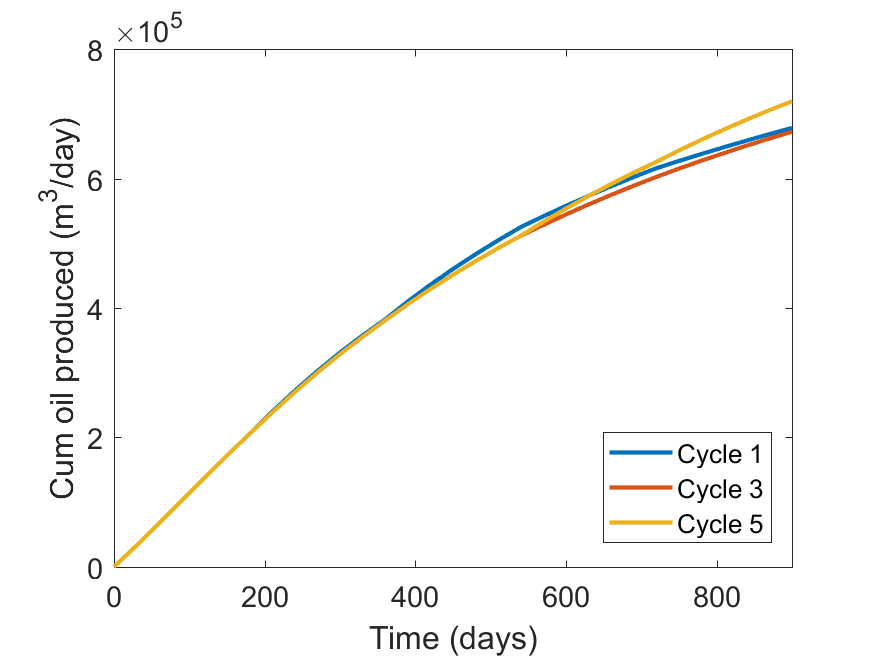}
\caption{Cumulative oil produced}
\label{fig:stepCOP_C}
\end{subfigure}
\begin{subfigure}{8.6cm}
\includegraphics[width=\textwidth]{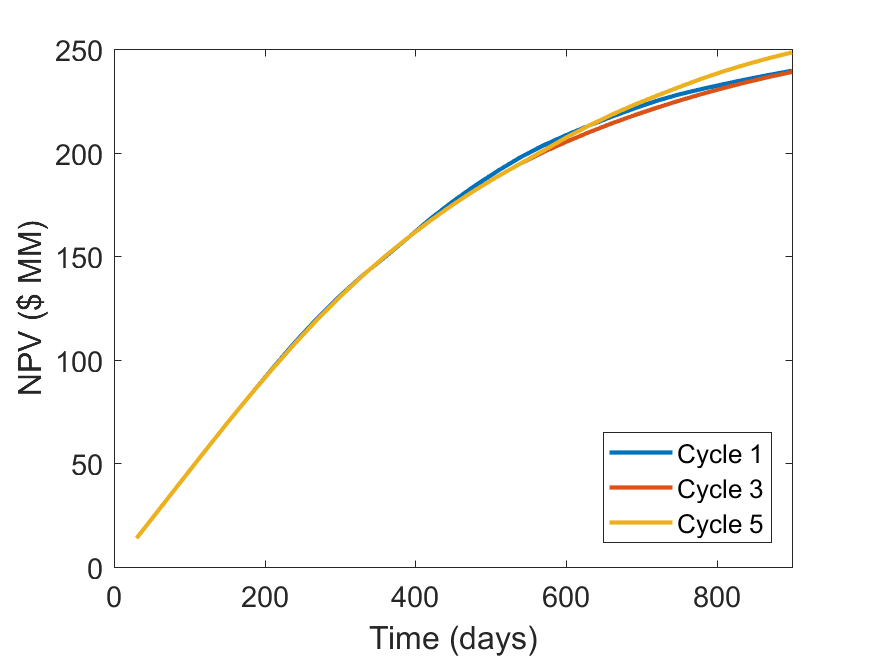}
\caption{NPV}
\label{fig:stepNPV_C}
\end{subfigure}
\caption{Optimum solutions found by CNN-RNN proxy at CLRM cycles~1, 3 and 5 for True model~C. The results shown are obtained by performing simulation using BHPs from proxy-based optimization at different CLRM cycles.} \label{fig:stepcompareC}
\end{figure*}

\begin{figure*}[htbp!]\centering
\begin{subfigure}{8.6cm}
\includegraphics[width=\textwidth]{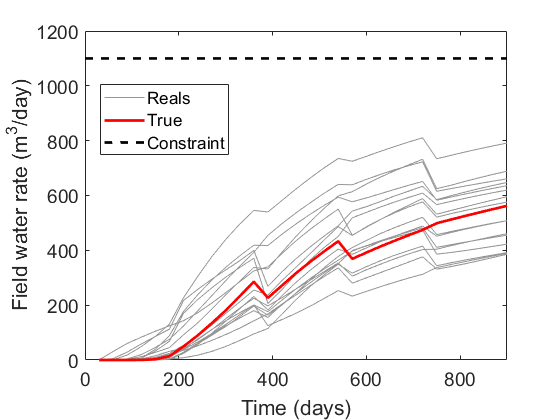}
\caption{Field water production rate at cycle~1} 
\end{subfigure}
\begin{subfigure}{8.6cm}
\includegraphics[width=\textwidth]{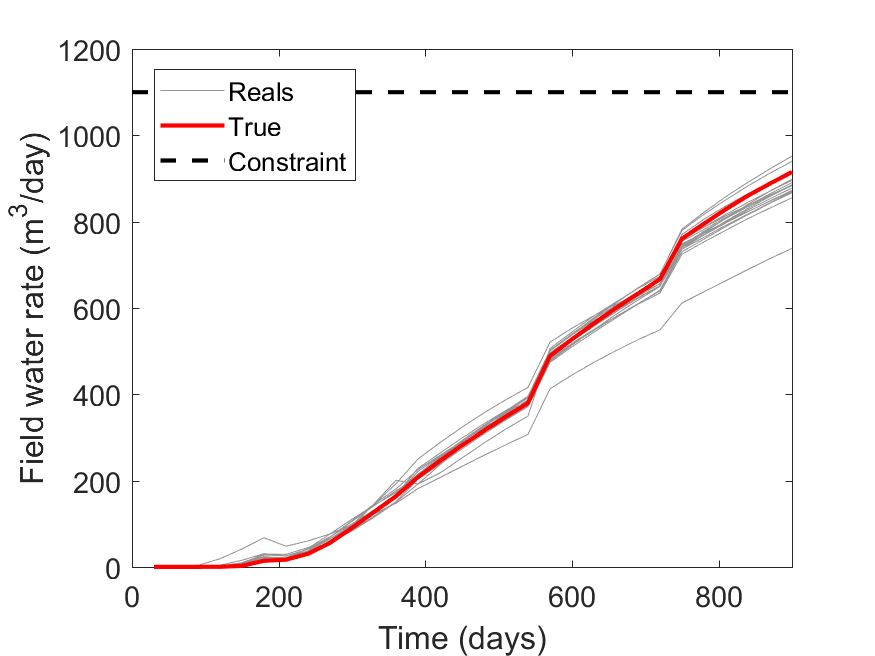}
\caption{Field water production rate at cycle~5} 
\end{subfigure}
\begin{subfigure}{8.6cm}
\includegraphics[width=\textwidth]{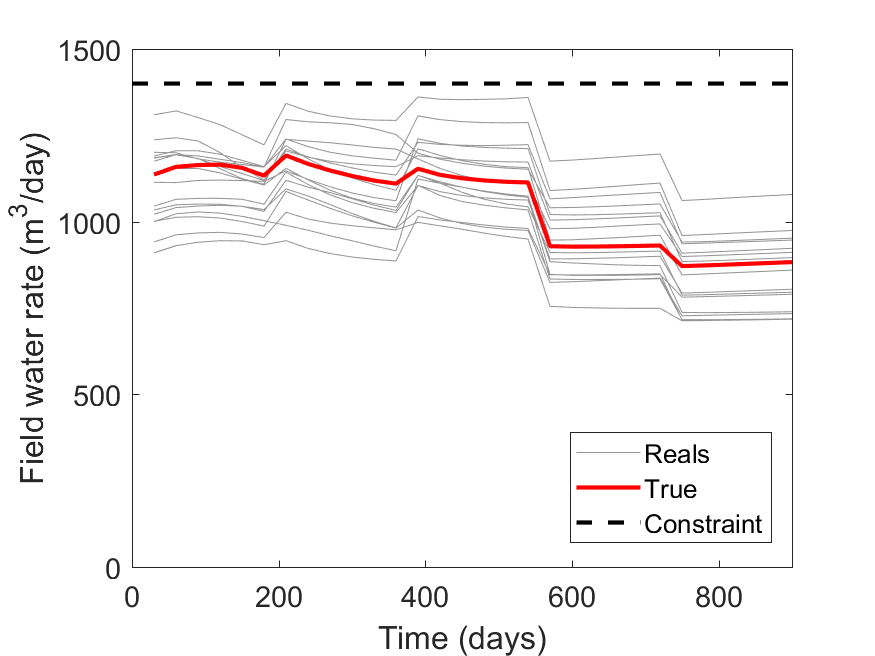}
\caption{Field water injection rate at cycle~1}
\end{subfigure}
\begin{subfigure}{8.6cm}
\includegraphics[width=\textwidth]{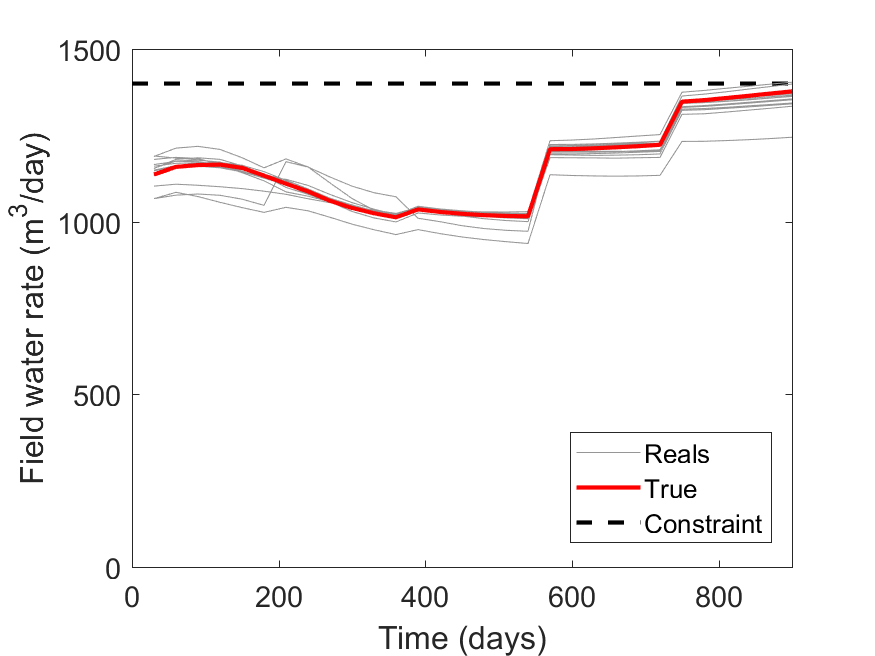}
\caption{Field water injection rate at cycle~5}
\end{subfigure}
\begin{subfigure}{8.6cm}
\includegraphics[width=\textwidth]{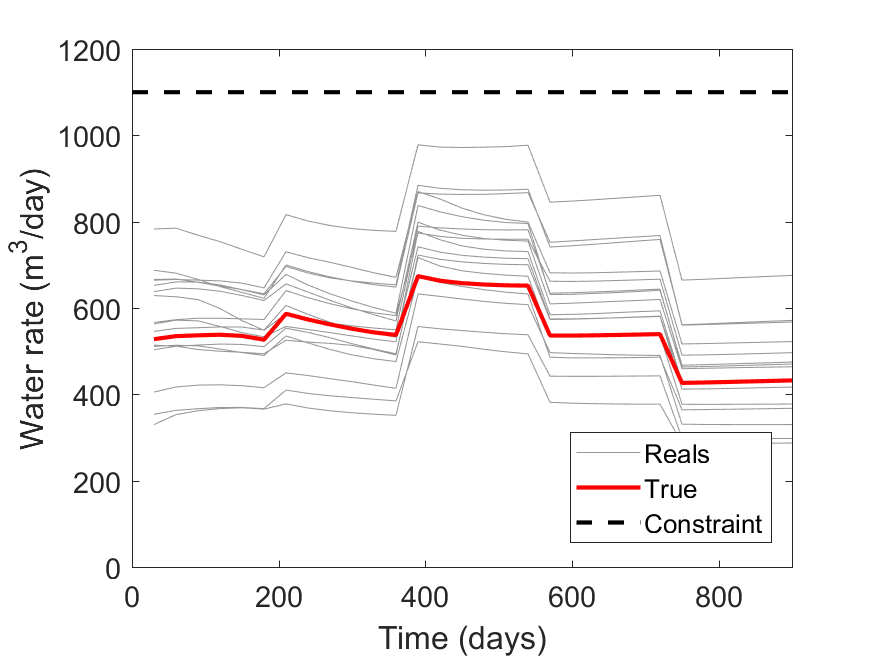}
\caption{INJ2 water rate at cycle~1}
\end{subfigure}
\begin{subfigure}{8.6cm}
\includegraphics[width=\textwidth]{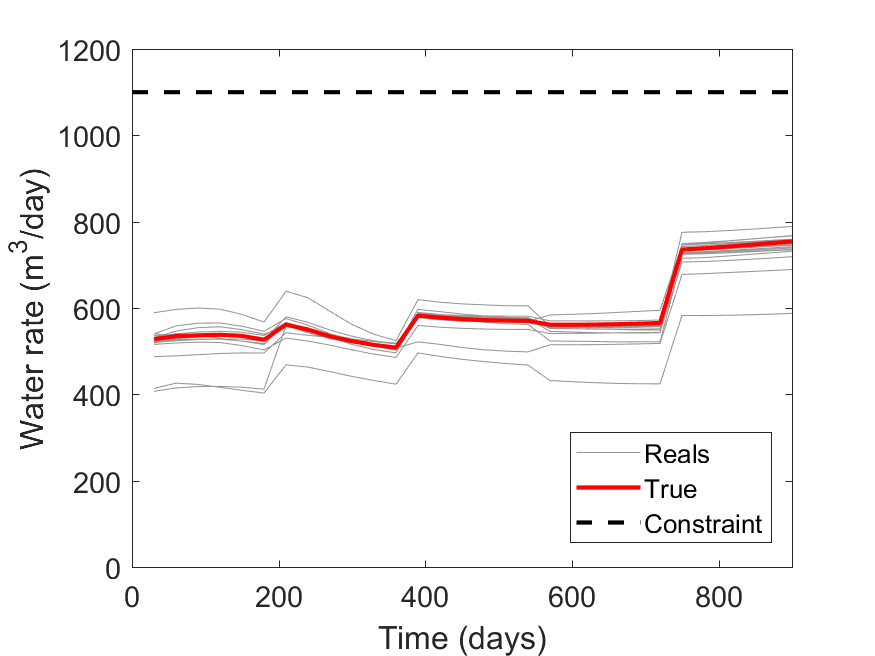}
\caption{INJ2 water rate at cycle~5}
\end{subfigure}
\caption{Optimized solutions at CLRM cycles~1 and 5 for True model~C. Results shown for $n_{ro}=16$ realizations (gray curves) and for true model. Nonlinear constraint limits shown as dashed lines.} \label{fig:constraintTrueC}
\end{figure*}

We now show an analogous set of results for True model~C. Less improvement in NPV from cycle~1 to cycle~5 is observed in this case than for True model~A (3.7\% versus 5.9\%). This case is interesting, however, because the true model NPV at cycle~1 is located at the edge of the NPV distribution (Figure~\ref{fig:CLRM_TC}). Figures~\ref{fig:stepcompareC} and \ref{fig:constraintTrueC} show field-wide responses and optimal rates at different CLRM cycles. We again see increased oil production from cycle~1 to cycle~5, facilitated by an increase in water injection. In Figure~\ref{fig:constraintTrueC}, less variation over the $n_{ro}$ realizations from cycle~1 to cycle~5 is again observed for the constraint quantities. As was the case for True model~A (Figure~\ref{fig:constraintTrueA}d), field water injection approaches the constraint limit at late time (Figure~\ref{fig:constraintTrueC}d). Again, the true model satisfies all constraints over the entire simulation time frame. This case illustrates that the proxy-based CLRM procedure is effective even for models that lie near the edge of the prior distribution.

Finally, we discuss the computational requirements associated with robust optimization and history matching in CLRM. Optimizations using the CNN-RNN proxy do not require any simulations during the optimization run itself, though the proxy must be trained initially and retrained after each data assimilation step. This entails 300 high-fidelity simulation runs for the initial training and 200 high-fidelity simulation runs for each retraining step. The number of simulation runs required to perform robust optimization in the proxy-based CLRM, for one true model, is thus $300 + 200 \times 4=1100$. 

If traditional CLRM is performed using high-fidelity simulation runs, the number of simulations required in the robust optimization steps is $5 \times 35 \times 30 \times 20=105,000$ (5~CLRM cycles, 35~PSO particles, 30~PSO iterations, 20~realizations). The speedup achieved using our proxy model, in terms of the number of high-fidelity simulation runs required for robust optimization, is thus about a factor of 100. Actual speedup is less since the proxy must be trained, which requires about 90~minutes for the initial training and 2--25~minutes for each retraining. This training time, however, is independent of the size of the model (though it will depend on the number of wells), so with large models it will be very small compared to the time required for robust optimization. Note also that the use of restarts could reduce the timings for robust optimization in traditional CLRM, since all runs use the same BHP schedules at previous CLRM cycles.

The computational requirements for history matching are the same in our framework as in traditional CLRM, since we apply a simulation-based RML procedure. The cost of this is, however, much less than that for robust optimization. Specifically, we perform 20 RML runs, with each run requiring about the equivalent of 110 simulations. The number of runs required for all history matching steps is thus about $4 \times 110 \times 20 = 8800$ (history matching is not performed at the first CLRM cycle). The total number of simulations required for traditional CLRM and proxy-based CLRM are $105,000 + 8800 = 113,800$ and $1100 + 8800 = 9900$, which corresponds to a speedup of 11.5 (not including training time). The computations associated with history matching could be reduced through use of an ensemble-based method, such as the ensemble smoother with multiple data assimilation \citep{EMERICK20133}, or through the use of a deep-learning proxy for history matching, such as those described in \citep{TANG2020109456, https://doi.org/10.1029/2018WR024638,JO2022109247}.

\clearpage
\section{Concluding Remarks}
\label{sec:conclusion}
In this work, we developed a CNN--RNN proxy to estimate well-by-well oil and water rates, as a function of time, over multiple realizations in an ensemble. The proxy, which extends the single-realization RNN-based procedure introduced in~\citep{Kim2021RNN}, accepts as input spatial permeability maps and well-by-well BHP schedules. The permeability field is initially processed by the CNN and is then fed into the LSTM RNN, which also accepts the time-varying well BHPs. The well rates provided by the network allow for the evaluation of nonlinear constraints, such as well or field-wide water injection or production rate, which must be treated in many production optimization settings.

The CNN--RNN proxy was tested for oil-water flow in 3D multi-Gaussian geomodels. Initial training involved 300 high-fidelity simulation runs, performed with different combinations of BHP profiles and permeability realizations. This training required about 1.5~hours using a Nvidia Tesla V100 GPU. The trained proxy provided reasonably accurate predictions of well-by-well oil and water rates, with 10th and 90th percentile overall errors of 5.1\% and 8.7\%. 

We then implemented the proxy model into a closed-loop reservoir management workflow. Robust optimization over an ensemble of geomodels was performed using the proxy model, within a PSO framework, in combination with a filter method to treat the nonlinear output constraints. History matching was accomplished using an adjoint-gradient-based randomized maximum likelihood method, with the geomodels parameterized using PCA. Proxy model retraining, which required only 2--25~minutes, was performed at each CLRM cycle. The proxy-based CLRM workflow was evaluated for five different (synthetic) `true' models. The overall procedure was shown to improve NPV from 2.7\% to 6.4\% relative to performing robust optimization with prior geomodels. Nonlinear constraint satisfaction was also demonstrated. The speedup achieved for the robust production optimization steps relative to a traditional (simulation-based) approach, quantified in terms of the number of simulations required, was about a factor of 100. Speedup factors are reduced if we account for training time or include the history matching computations (which are performed the same way in both the proxy-based and traditional CLRM workflows).

There are a number of directions that should be considered in future work. The overall CLRM workflow could be accelerated by incorporating proxy treatments for the history matching process. Many proxy models for history matching have been suggested, including those in \citep{TANG2020109456, https://doi.org/10.1029/2018WR024638,JO2022109247}, and these could be tested in this setting. It may also be useful to incorporate geomodel parameterization procedures applicable for non-Gaussian models (e.g., CNN--PCA, generative adversarial networks) to treat cases involving channelized systems. 
It is also of interest to test the CNN--RNN proxy for a range of more complex simulation setups, such as compositional systems and geomodels described on unstructured grids. Finally, testing of some or all components of the workflow should be performed for a range of field cases.

\begin{acknowledgements}
We are grateful to the Stanford Center for Computational Earth \& Environmental Sciences for providing computational resources. We thank Oleg Volkov for his assistance with the Stanford Unified Optimization Framework and Dylan Crain for his help with 3D model generation.
\end{acknowledgements}

\small {\noindent {\bf Funding information} The authors received financial support from the industrial affiliates of the Stanford Smart Fields Consortium.}

\small {\noindent {\bf Competing interests} We declare that this research was conducted in the absence of any commercial or financial relationships that could be construed as a potential conflict of interest.}

\small {\noindent {\bf Data availability} Inquiries regarding the data should be directed to Yong Do Kim, ykim07@stanford.edu.}

\bibliographystyle{elsarticle-num-names} 
\bibliography{ref}

\begin{thebibliography}{45}
\expandafter\ifx\csname natexlab\endcsname\relax\def\natexlab#1{#1}\fi
\providecommand{\url}[1]{\texttt{#1}}
\providecommand{\href}[2]{#2}
\providecommand{\path}[1]{#1}
\providecommand{\DOIprefix}{doi:}
\providecommand{\ArXivprefix}{arXiv:}
\providecommand{\URLprefix}{URL: }
\providecommand{\Pubmedprefix}{pmid:}
\providecommand{\doi}[1]{\href{http://dx.doi.org/#1}{\path{#1}}}
\providecommand{\Pubmed}[1]{\href{pmid:#1}{\path{#1}}}
\providecommand{\bibinfo}[2]{#2}
\ifx\xfnm\relax \def\xfnm[#1]{\unskip,\space#1}\fi
\bibitem[{Kim and Durlofsky(2021)}]{Kim2021RNN}
\bibinfo{author}{Y.~D. Kim}, \bibinfo{author}{L.~J. Durlofsky},
\newblock \bibinfo{title}{{A recurrent neural network–based proxy model for
  well-control optimization with nonlinear output constraints}},
\newblock \bibinfo{journal}{SPE Journal} \bibinfo{volume}{26}
  (\bibinfo{year}{2021}) \bibinfo{pages}{1837--1857}.
\bibitem[{Møyner et~al.(2015)Møyner, Krogstad, and Lie}]{moyner2015}
\bibinfo{author}{O.~Møyner}, \bibinfo{author}{S.~Krogstad},
  \bibinfo{author}{K.-A. Lie},
\newblock \bibinfo{title}{The application of flow diagnostics for reservoir
  management},
\newblock \bibinfo{journal}{SPE Journal} \bibinfo{volume}{20}
  (\bibinfo{year}{2015}) \bibinfo{pages}{306--323}.
\bibitem[{Rodríguez~Torrado et~al.(2015)Rodríguez~Torrado,
  Echeverría-Ciaurri, Mello, and Embid~Droz}]{rodriguez_torrado2015}
\bibinfo{author}{R.~Rodríguez~Torrado},
  \bibinfo{author}{D.~Echeverría-Ciaurri}, \bibinfo{author}{U.~Mello},
  \bibinfo{author}{S.~Embid~Droz},
\newblock \bibinfo{title}{Opening new opportunities with fast
  reservoir-performance evaluation under uncertainty: Brugge field case study},
\newblock \bibinfo{journal}{SPE Economics and Management} \bibinfo{volume}{7}
  (\bibinfo{year}{2015}) \bibinfo{pages}{84--99}.
\bibitem[{{de Brito} and Durlofsky(2020)}]{deBrito2020}
\bibinfo{author}{D.~U. {de Brito}}, \bibinfo{author}{L.~J. Durlofsky},
\newblock \bibinfo{title}{Well control optimization using a two-step surrogate
  treatment},
\newblock \bibinfo{journal}{Journal of Petroleum Science and Engineering}
  \bibinfo{volume}{187} (\bibinfo{year}{2020}) \bibinfo{pages}{106565}.
\bibitem[{Jansen and Durlofsky(2017)}]{Jansen2017UseOR}
\bibinfo{author}{J.~D. Jansen}, \bibinfo{author}{L.~J. Durlofsky},
\newblock \bibinfo{title}{Use of reduced-order models in well control
  optimization},
\newblock \bibinfo{journal}{Optimization and Engineering} \bibinfo{volume}{18}
  (\bibinfo{year}{2017}) \bibinfo{pages}{105--132}.
\bibitem[{Guo and Reynolds(2018)}]{Guo2018RobustLP}
\bibinfo{author}{Z.~Guo}, \bibinfo{author}{A.~C. Reynolds},
\newblock \bibinfo{title}{Robust life-cycle production optimization with a
  support-vector-regression proxy},
\newblock \bibinfo{journal}{SPE Journal} \bibinfo{volume}{23}
  (\bibinfo{year}{2018}) \bibinfo{pages}{2409--2427}.
\bibitem[{Chen et~al.(2020)Chen, Zhang, Zhang, Xue, Ji, Yao, Yao, and
  Yang}]{Chen2020GlobalAL}
\bibinfo{author}{G.~Chen}, \bibinfo{author}{K.~Zhang},
  \bibinfo{author}{L.~Zhang}, \bibinfo{author}{X.~Xue},
  \bibinfo{author}{D.~Ji}, \bibinfo{author}{C.~Yao}, \bibinfo{author}{J.~Yao},
  \bibinfo{author}{Y.~Yang},
\newblock \bibinfo{title}{Global and local surrogate-model-assisted
  differential evolution for waterflooding production optimization},
\newblock \bibinfo{journal}{SPE Journal} \bibinfo{volume}{25}
  (\bibinfo{year}{2020}) \bibinfo{pages}{105--118}.
\bibitem[{Zangl et~al.(2006)Zangl, Graf, and Al-Kinani}]{Zangl2006}
\bibinfo{author}{G.~Zangl}, \bibinfo{author}{T.~Graf},
  \bibinfo{author}{A.~Al-Kinani},
\newblock \bibinfo{title}{Proxy modeling in production optimization}
  (\bibinfo{year}{2006}). \bibinfo{note}{Paper presented at the SPE
  Europec/EAGE Annual Conference and Exhibition, Vienna, Austria, June 2006.}
\bibitem[{Golzari et~al.(2015)Golzari, Sefat, and
  Jamshidi}]{Golzari2015DevelopmentOA}
\bibinfo{author}{A.~Golzari}, \bibinfo{author}{M.~H. Sefat},
  \bibinfo{author}{S.~Jamshidi},
\newblock \bibinfo{title}{Development of an adaptive surrogate model for
  production optimization},
\newblock \bibinfo{journal}{Journal of Petroleum Science and Engineering}
  \bibinfo{volume}{133} (\bibinfo{year}{2015}) \bibinfo{pages}{677--688}.
\bibitem[{Zhao et~al.(2020)Zhao, Zhang, Chen, Zhao, Yao, Sun, Huang, and
  Yao}]{ZHAO2020107192}
\bibinfo{author}{M.~Zhao}, \bibinfo{author}{K.~Zhang},
  \bibinfo{author}{G.~Chen}, \bibinfo{author}{X.~Zhao},
  \bibinfo{author}{C.~Yao}, \bibinfo{author}{H.~Sun},
  \bibinfo{author}{Z.~Huang}, \bibinfo{author}{J.~Yao},
\newblock \bibinfo{title}{A surrogate-assisted multi-objective evolutionary
  algorithm with dimension-reduction for production optimization},
\newblock \bibinfo{journal}{Journal of Petroleum Science and Engineering}
  \bibinfo{volume}{192} (\bibinfo{year}{2020}) \bibinfo{pages}{107192}.
\bibitem[{Zhang and Sheng(2021)}]{10.2118/206755-PA}
\bibinfo{author}{H.~Zhang}, \bibinfo{author}{J.~J. Sheng},
\newblock \bibinfo{title}{{Surrogate-assisted multiobjective optimization of a
  hydraulically fractured well in a naturally fractured shale reservoir with
  geological uncertainty}},
\newblock \bibinfo{journal}{SPE Journal}  (\bibinfo{year}{2021})
  \bibinfo{pages}{1--22}.
\bibitem[{Petvipusit et~al.(2014)Petvipusit, Elsheikh, Laforce, King, and
  Blunt}]{Petvipusit2014}
\bibinfo{author}{K.~R. Petvipusit}, \bibinfo{author}{A.~H. Elsheikh},
  \bibinfo{author}{T.~C. Laforce}, \bibinfo{author}{P.~R. King},
  \bibinfo{author}{M.~J. Blunt},
\newblock \bibinfo{title}{Robust optimisation of {C}{O}$_2$ sequestration
  strategies under geological uncertainty using adaptive sparse grid
  surrogates},
\newblock \bibinfo{journal}{Computational Geosciences} \bibinfo{volume}{18}
  (\bibinfo{year}{2014}) \bibinfo{pages}{763--778}.
\bibitem[{Babaei and Pan(2016)}]{Babaei2016}
\bibinfo{author}{M.~Babaei}, \bibinfo{author}{I.~Pan},
\newblock \bibinfo{title}{Performance comparison of several response surface
  surrogate models and ensemble methods for water injection optimization under
  uncertainty},
\newblock \bibinfo{journal}{Computers \& Geosciences} \bibinfo{volume}{91}
  (\bibinfo{year}{2016}).
\bibitem[{Kim et~al.(2020)Kim, Yang, and Choe}]{KIM2020107424}
\bibinfo{author}{J.~Kim}, \bibinfo{author}{H.~Yang}, \bibinfo{author}{J.~Choe},
\newblock \bibinfo{title}{Robust optimization of the locations and types of
  multiple wells using {CNN} based proxy models},
\newblock \bibinfo{journal}{Journal of Petroleum Science and Engineering}
  \bibinfo{volume}{193} (\bibinfo{year}{2020}) \bibinfo{pages}{107424}.
\bibitem[{Wang et~al.(2022)Wang, Chang, Zhang, Xue, and Chen}]{WANG2022109545}
\bibinfo{author}{N.~Wang}, \bibinfo{author}{H.~Chang},
  \bibinfo{author}{D.~Zhang}, \bibinfo{author}{L.~Xue},
  \bibinfo{author}{Y.~Chen},
\newblock \bibinfo{title}{Efficient well placement optimization based on
  theory-guided convolutional neural network},
\newblock \bibinfo{journal}{Journal of Petroleum Science and Engineering}
  \bibinfo{volume}{208} (\bibinfo{year}{2022}) \bibinfo{pages}{109545}.
\bibitem[{Nwachukwu et~al.(2018)Nwachukwu, Jeong, Sun, Pyrcz, and
  Lake}]{Nwachukwu2018}
\bibinfo{author}{A.~Nwachukwu}, \bibinfo{author}{H.~Jeong},
  \bibinfo{author}{A.~Sun}, \bibinfo{author}{M.~Pyrcz}, \bibinfo{author}{L.~W.
  Lake},
\newblock \bibinfo{title}{Machine learning-based optimization of well locations
  and {WAG} parameters under geologic uncertainty}  (\bibinfo{year}{2018}).
  \bibinfo{note}{Paper presented at the SPE Improved Oil Recovery Conference,
  Tulsa, Oklahoma, April 2018.}
\bibitem[{Jansen et~al.(2009)Jansen, Brouwer, and Douma}]{Jansen2009}
\bibinfo{author}{J.~D. Jansen}, \bibinfo{author}{R.~Brouwer},
  \bibinfo{author}{S.~G. Douma},
\newblock \bibinfo{title}{Closed loop reservoir management}
  (\bibinfo{year}{2009}). \bibinfo{note}{Paper presented at the SPE Reservoir
  Simulation Symposium, The Woodlands, Texas, February 2009.}
\bibitem[{Sarma et~al.(2006)Sarma, Durlofsky, Aziz, and Chen}]{Sarma2006}
\bibinfo{author}{P.~Sarma}, \bibinfo{author}{L.~J. Durlofsky},
  \bibinfo{author}{K.~Aziz}, \bibinfo{author}{W.~H. Chen},
\newblock \bibinfo{title}{Efficient real-time reservoir management using
  adjoint-based optimal control and model updating},
\newblock \bibinfo{journal}{Computational Geosciences} \bibinfo{volume}{10}
  (\bibinfo{year}{2006}) \bibinfo{pages}{3--36}.
\bibitem[{Wang et~al.(2009)Wang, Li, and Reynolds}]{10.2118/109805-PA}
\bibinfo{author}{C.~Wang}, \bibinfo{author}{G.~Li}, \bibinfo{author}{A.~C.
  Reynolds},
\newblock \bibinfo{title}{{Production optimization in closed-loop reservoir
  management}},
\newblock \bibinfo{journal}{SPE Journal} \bibinfo{volume}{14}
  (\bibinfo{year}{2009}) \bibinfo{pages}{506--523}.
\bibitem[{Fonseca et~al.(2015{\natexlab{a}})Fonseca, Leeuwenburgh, Van~den Hof,
  and Jansen}]{10.2118/163657-PA}
\bibinfo{author}{R.~M. Fonseca}, \bibinfo{author}{O.~Leeuwenburgh},
  \bibinfo{author}{P.~M. Van~den Hof}, \bibinfo{author}{J.~D. Jansen},
\newblock \bibinfo{title}{{Improving the ensemble-optimization method through
  covariance-matrix adaptation}},
\newblock \bibinfo{journal}{SPE Journal} \bibinfo{volume}{20}
  (\bibinfo{year}{2015}{\natexlab{a}}) \bibinfo{pages}{155--168}.
\bibitem[{Fonseca et~al.(2015{\natexlab{b}})Fonseca, Leeuwenburgh, Della~Rossa,
  Van~den Hof, and Jansen}]{10.2118/173268-PA}
\bibinfo{author}{R.~M. Fonseca}, \bibinfo{author}{O.~Leeuwenburgh},
  \bibinfo{author}{E.~Della~Rossa}, \bibinfo{author}{P.~M. Van~den Hof},
  \bibinfo{author}{J.~D. Jansen},
\newblock \bibinfo{title}{{Ensemble-based multiobjective optimization of on/off
  control devices under geological uncertainty}},
\newblock \bibinfo{journal}{SPE Reservoir Evaluation \& Engineering}
  \bibinfo{volume}{18} (\bibinfo{year}{2015}{\natexlab{b}})
  \bibinfo{pages}{554--563}.
\bibitem[{Ramaswamy et~al.(2020)Ramaswamy, Fonseca, Leeuwenburgh, Siraj, and
  Van~den Hof}]{Ramaswamy2020}
\bibinfo{author}{K.~R. Ramaswamy}, \bibinfo{author}{R.~M. Fonseca},
  \bibinfo{author}{O.~Leeuwenburgh}, \bibinfo{author}{M.~M. Siraj},
  \bibinfo{author}{P.~M.~J. Van~den Hof},
\newblock \bibinfo{title}{Improved sampling strategies for ensemble-based
  optimization},
\newblock \bibinfo{journal}{Computational Geosciences} \bibinfo{volume}{24}
  (\bibinfo{year}{2020}) \bibinfo{pages}{1057--1069}.
\bibitem[{Jiang et~al.(2020)Jiang, Sun, and Durlofsky}]{Jiang2020}
\bibinfo{author}{S.~Jiang}, \bibinfo{author}{W.~Sun}, \bibinfo{author}{L.~J.
  Durlofsky},
\newblock \bibinfo{title}{A data-space inversion procedure for well control
  optimization and closed-loop reservoir management},
\newblock \bibinfo{journal}{Computational Geosciences} \bibinfo{volume}{24}
  (\bibinfo{year}{2020}) \bibinfo{pages}{361--379}.
\bibitem[{Deng and Pan(2020)}]{10.2118/200862-PA}
\bibinfo{author}{L.~Deng}, \bibinfo{author}{Y.~Pan},
\newblock \bibinfo{title}{{Machine-learning-assisted closed-loop reservoir
  management using echo state network for mature fields under waterflood}},
\newblock \bibinfo{journal}{SPE Reservoir Evaluation \& Engineering}
  \bibinfo{volume}{23} (\bibinfo{year}{2020}) \bibinfo{pages}{1298--1313}.
\bibitem[{Schmidhuber(2015)}]{SCHMIDHUBER201585}
\bibinfo{author}{J.~Schmidhuber},
\newblock \bibinfo{title}{Deep learning in neural networks: An overview},
\newblock \bibinfo{journal}{Neural Networks} \bibinfo{volume}{61}
  (\bibinfo{year}{2015}) \bibinfo{pages}{85 -- 117}.
\bibitem[{LeCun et~al.(2015)LeCun, Bengio, and Hinton}]{LeCun2015}
\bibinfo{author}{Y.~LeCun}, \bibinfo{author}{Y.~Bengio},
  \bibinfo{author}{G.~Hinton},
\newblock \bibinfo{title}{Deep learning},
\newblock \bibinfo{journal}{Nature} \bibinfo{volume}{521}
  (\bibinfo{year}{2015}) \bibinfo{pages}{436--444}.
\bibitem[{Karpathy and Fei-Fei(2015)}]{Karpathy_2015_CVPR}
\bibinfo{author}{A.~Karpathy}, \bibinfo{author}{L.~Fei-Fei},
\newblock \bibinfo{title}{Deep visual-semantic alignments for generating image
  descriptions},
\newblock in: \bibinfo{booktitle}{Proceedings of the IEEE Conference on
  Computer Vision and Pattern Recognition (CVPR)}, \bibinfo{year}{2015}.
\bibitem[{Tang et~al.(2020)Tang, Liu, and Durlofsky}]{TANG2020109456}
\bibinfo{author}{M.~Tang}, \bibinfo{author}{Y.~Liu}, \bibinfo{author}{L.~J.
  Durlofsky},
\newblock \bibinfo{title}{A deep-learning-based surrogate model for data
  assimilation in dynamic subsurface flow problems},
\newblock \bibinfo{journal}{Journal of Computational Physics}
  \bibinfo{volume}{413} (\bibinfo{year}{2020}) \bibinfo{pages}{109456}.
\bibitem[{Gers et~al.(2000)Gers, Schmidhuber, and Cummins}]{Gers2000LearningTF}
\bibinfo{author}{F.~A. Gers}, \bibinfo{author}{J.~Schmidhuber},
  \bibinfo{author}{F.~A. Cummins},
\newblock \bibinfo{title}{Learning to forget: Continual prediction with
  {LSTM}},
\newblock \bibinfo{journal}{Neural Computation} \bibinfo{volume}{12}
  (\bibinfo{year}{2000}) \bibinfo{pages}{2451--2471}.
\bibitem[{Isebor et~al.(2014)Isebor, Durlofsky, and
  Echeverría~Ciaurri}]{Isebor2014}
\bibinfo{author}{O.~J. Isebor}, \bibinfo{author}{L.~J. Durlofsky},
  \bibinfo{author}{D.~Echeverría~Ciaurri},
\newblock \bibinfo{title}{A derivative-free methodology with local and global
  search for the constrained joint optimization of well locations and
  controls},
\newblock \bibinfo{journal}{Computational Geosciences} \bibinfo{volume}{18}
  (\bibinfo{year}{2014}) \bibinfo{pages}{463--482}.
\bibitem[{Abadi et~al.(2015)Abadi, Agarwal, Barham, Brevdo, Chen, Citro,
  Corrado, Davis, Dean, Devin, Ghemawat, Goodfellow, Harp, Irving, Isard, Jia,
  Jozefowicz, Kaiser, Kudlur, Levenberg, Man\'{e}, Monga, Moore, Murray, Olah,
  Schuster, Shlens, Steiner, Sutskever, Talwar, Tucker, Vanhoucke, Vasudevan,
  Vi\'{e}gas, Vinyals, Warden, Wattenberg, Wicke, Yu, and
  Zheng}]{tensorflow2015-whitepaper}
\bibinfo{author}{M.~Abadi}, \bibinfo{author}{A.~Agarwal},
  \bibinfo{author}{P.~Barham}, \bibinfo{author}{E.~Brevdo},
  \bibinfo{author}{Z.~Chen}, \bibinfo{author}{C.~Citro}, \bibinfo{author}{G.~S.
  Corrado}, \bibinfo{author}{A.~Davis}, \bibinfo{author}{J.~Dean},
  \bibinfo{author}{M.~Devin}, \bibinfo{author}{S.~Ghemawat},
  \bibinfo{author}{I.~Goodfellow}, \bibinfo{author}{A.~Harp},
  \bibinfo{author}{G.~Irving}, \bibinfo{author}{M.~Isard},
  \bibinfo{author}{Y.~Jia}, \bibinfo{author}{R.~Jozefowicz},
  \bibinfo{author}{L.~Kaiser}, \bibinfo{author}{M.~Kudlur},
  \bibinfo{author}{J.~Levenberg}, \bibinfo{author}{D.~Man\'{e}},
  \bibinfo{author}{R.~Monga}, \bibinfo{author}{S.~Moore},
  \bibinfo{author}{D.~Murray}, \bibinfo{author}{C.~Olah},
  \bibinfo{author}{M.~Schuster}, \bibinfo{author}{J.~Shlens},
  \bibinfo{author}{B.~Steiner}, \bibinfo{author}{I.~Sutskever},
  \bibinfo{author}{K.~Talwar}, \bibinfo{author}{P.~Tucker},
  \bibinfo{author}{V.~Vanhoucke}, \bibinfo{author}{V.~Vasudevan},
  \bibinfo{author}{F.~Vi\'{e}gas}, \bibinfo{author}{O.~Vinyals},
  \bibinfo{author}{P.~Warden}, \bibinfo{author}{M.~Wattenberg},
  \bibinfo{author}{M.~Wicke}, \bibinfo{author}{Y.~Yu},
  \bibinfo{author}{X.~Zheng}, \bibinfo{title}{{TensorFlow}: Large-scale machine
  learning on heterogeneous systems}, \bibinfo{year}{2015}.
\bibitem[{Ioffe and Szegedy(2015)}]{Sergey2015}
\bibinfo{author}{S.~Ioffe}, \bibinfo{author}{C.~Szegedy},
\newblock \bibinfo{title}{Batch normalization: Accelerating deep network
  training by reducing internal covariate shift},
\newblock \bibinfo{journal}{Proceedings of the 32nd International Conference on
  International Conference on Machine Learning}  (\bibinfo{year}{2015})
  \bibinfo{pages}{448--456}.
\bibitem[{G\'eron(2019)}]{Geron}
\bibinfo{author}{A.~G\'eron}, \bibinfo{title}{Hands-On Machine Learning with
  Scikit-Learn and TensorFlow: Concepts, Tools, and Techniques to Build
  Intelligent Systems}, \bibinfo{publisher}{O'Reilly Media, Inc},
  \bibinfo{address}{Sebastopol}, \bibinfo{year}{2019}.
\bibitem[{Zhou(2012)}]{zhouThesis}
\bibinfo{author}{Y.~Zhou}, \bibinfo{title}{Parallel {G}eneral-purpose
  {R}eservoir {S}imulation with {C}oupled {R}eservoir {M}odels and
  {M}ultisegment {W}ells}, Ph.D. thesis, Stanford University,
  \bibinfo{year}{2012}.
\bibitem[{Kingma and Ba(2014)}]{Kingma2014}
\bibinfo{author}{D.~P. Kingma}, \bibinfo{author}{J.~Ba},
\newblock \bibinfo{title}{Adam: A method for stochastic optimization}
  (\bibinfo{year}{2014}). \bibinfo{note}{ArXiv preprint arXiv:1412.6980}.
\bibitem[{Chen et~al.(2012)Chen, Li, and Reynolds}]{10.2118/141314-PA}
\bibinfo{author}{C.~Chen}, \bibinfo{author}{G.~Li}, \bibinfo{author}{A.~C.
  Reynolds},
\newblock \bibinfo{title}{{Robust constrained optimization of short- and
  long-term net present value for closed-loop reservoir management}},
\newblock \bibinfo{journal}{SPE Journal} \bibinfo{volume}{17}
  (\bibinfo{year}{2012}) \bibinfo{pages}{849--864}.
\bibitem[{Hanssen et~al.(2015)Hanssen, Foss, and Teixeira}]{HANSSEN201562}
\bibinfo{author}{K.~G. Hanssen}, \bibinfo{author}{B.~Foss},
  \bibinfo{author}{A.~Teixeira},
\newblock \bibinfo{title}{Production optimization under uncertainty with
  constraint handling},
\newblock \bibinfo{journal}{IFAC-PapersOnLine} \bibinfo{volume}{48}
  (\bibinfo{year}{2015}) \bibinfo{pages}{62--67}. \bibinfo{note}{2nd IFAC
  Workshop on Automatic Control in Offshore Oil and Gas Production OOGP 2015}.
\bibitem[{Aliyev and Durlofsky(2017)}]{Aliyev2017}
\bibinfo{author}{E.~Aliyev}, \bibinfo{author}{L.~J. Durlofsky},
\newblock \bibinfo{title}{{Multilevel field development optimization under
  uncertainty using a sequence of upscaled models}},
\newblock \bibinfo{journal}{Mathematical Geosciences} \bibinfo{volume}{49}
  (\bibinfo{year}{2017}) \bibinfo{pages}{307--339}.
\bibitem[{Liu and Durlofsky(2021)}]{LIU2021104676}
\bibinfo{author}{Y.~Liu}, \bibinfo{author}{L.~J. Durlofsky},
\newblock \bibinfo{title}{3{D CNN-PCA}: A deep-learning-based parameterization
  for complex geomodels},
\newblock \bibinfo{journal}{Computers \& Geosciences} \bibinfo{volume}{148}
  (\bibinfo{year}{2021}) \bibinfo{pages}{104676}.
\bibitem[{Remy et~al.(2009)Remy, Boucher, and Wu}]{SGeMs}
\bibinfo{author}{N.~Remy}, \bibinfo{author}{A.~Boucher},
  \bibinfo{author}{J.~Wu}, \bibinfo{title}{Applied Geostatistics with SGeMS: A
  User's Guide}, \bibinfo{publisher}{Cambridge University Press},
  \bibinfo{address}{New York}, \bibinfo{year}{2009}.
\bibitem[{Vo and Durlofsky(2015)}]{Vo2015}
\bibinfo{author}{H.~X. Vo}, \bibinfo{author}{L.~J. Durlofsky},
\newblock \bibinfo{title}{Data assimilation and uncertainty assessment for
  complex geological models using a new {PCA}-based parameterization},
\newblock \bibinfo{journal}{Computational Geosciences} \bibinfo{volume}{19}
  (\bibinfo{year}{2015}) \bibinfo{pages}{747--767}.
\bibitem[{Tang et~al.(2021)Tang, Liu, and Durlofsky}]{TANG2021113636}
\bibinfo{author}{M.~Tang}, \bibinfo{author}{Y.~Liu}, \bibinfo{author}{L.~J.
  Durlofsky},
\newblock \bibinfo{title}{Deep-learning-based surrogate flow modeling and
  geological parameterization for data assimilation in 3{D} subsurface flow},
\newblock \bibinfo{journal}{Computer Methods in Applied Mechanics and
  Engineering} \bibinfo{volume}{376} (\bibinfo{year}{2021})
  \bibinfo{pages}{113636}.
\bibitem[{Emerick and Reynolds(2013)}]{EMERICK20133}
\bibinfo{author}{A.~A. Emerick}, \bibinfo{author}{A.~C. Reynolds},
\newblock \bibinfo{title}{Ensemble smoother with multiple data assimilation},
\newblock \bibinfo{journal}{Computers \& Geosciences} \bibinfo{volume}{55}
  (\bibinfo{year}{2013}) \bibinfo{pages}{3--15}.
\bibitem[{Mo et~al.(2019)Mo, Zabaras, Shi, and
  Wu}]{https://doi.org/10.1029/2018WR024638}
\bibinfo{author}{S.~Mo}, \bibinfo{author}{N.~Zabaras},
  \bibinfo{author}{X.~Shi}, \bibinfo{author}{J.~Wu},
\newblock \bibinfo{title}{Deep autoregressive neural networks for
  high-dimensional inverse problems in groundwater contaminant source
  identification},
\newblock \bibinfo{journal}{Water Resources Research} \bibinfo{volume}{55}
  (\bibinfo{year}{2019}) \bibinfo{pages}{3856--3881}.
\bibitem[{Jo et~al.(2022)Jo, Jeong, Min, Park, Kim, Kwon, and
  Sun}]{JO2022109247}
\bibinfo{author}{S.~Jo}, \bibinfo{author}{H.~Jeong}, \bibinfo{author}{B.~Min},
  \bibinfo{author}{C.~Park}, \bibinfo{author}{Y.~Kim},
  \bibinfo{author}{S.~Kwon}, \bibinfo{author}{A.~Sun},
\newblock \bibinfo{title}{Efficient deep-learning-based history matching for
  fluvial channel reservoirs},
\newblock \bibinfo{journal}{Journal of Petroleum Science and Engineering}
  \bibinfo{volume}{208} (\bibinfo{year}{2022}) \bibinfo{pages}{109247}.

\end{thebibliography}

\end{document}